\documentclass[onecolumn]{elsarticle}
\usepackage[T1]{fontenc}
\usepackage[utf8]{inputenc}
\usepackage{graphicx}
\usepackage{cellspace}
\usepackage{amsmath}
\usepackage{subcaption}
\usepackage{adjustbox}
\usepackage{multirow}
\usepackage{algorithm, algpseudocode}

\usepackage{natbib}

\usepackage{hyperref}

\journal{Journal of Expert Systems with Applications}

\begin{document}

\begin{frontmatter}

\title{Learning Financial Asset-Specific Trading Rules via Deep Reinforcement Learning}

\author[mymainaddress,mysecondaryaddress]{Mehran Taghian}
\ead{mehrantaghian@aut.ac.ir}
\author[mymainaddress,mysecondaryaddress]{Ahmad Asadi}
\ead{ahmad.asadi@aut.ac.ir}
\author[mymainaddress,mysecondaryaddress]{Reza Safabakhsh\corref{mycorrespondingauthor}}
\cortext[mycorrespondingauthor]{Corresponding author}
\ead{safa@aut.ac.ir}




\address[mymainaddress]{Deep Learning Lab, Computer Engineering Department}
\address[mysecondaryaddress]{Amirkabir University of Technology, Hafez St., Tehran, Iran.}

\begin{abstract}
    	Generating asset-specific trading signals based on the financial conditions of the assets is one of the challenging problems in automated trading. Various asset trading rules are proposed experimentally based on different technical analysis techniques. However, these kind of trading strategies are profitable, extracting new asset-specific trading rules from vast historical data to increase total return and decrease the risk of portfolios is difficult for human experts. Recently, various deep reinforcement learning (DRL) methods are employed to learn the new trading rules for each asset. In this paper, a novel DRL model with various feature extraction modules is proposed. The effect of different input representations on the performance of the models is investigated and the performance of DRL-based models in different markets and asset situations is studied. The proposed model in this work outperformed the other state-of-the-art models in learning single asset-specific trading rules and obtained a total return of almost 262\% in two years on a specific asset while the best state-of-the-art model get 78\% on the same asset in the same time period. 
\end{abstract}

\begin{keyword}
	Reinforcement learning\sep
	Deep Q-learning\sep
	Single Stock trading\sep
	Trading strategy\sep
\end{keyword}

\end{frontmatter}

\section{Introduction}
\label{sec:intro}

In the face of growing demand by the investors for automated trading of financial assets in different markets, designing a model capable of learning appropriate trading rules and strategies attracted the attention of researchers from finance and artificial intelligence. The power of computers to process a vast historical data of the financial assets and to model price fluctuations and learn fitted trading rules, raises the idea of designing applications that can generate appropriate trading signals on a specific financial asset at each time step. 

Following the idea of replacing human traders with computer programs, a lot of research is conducted. The first issue to be addressed was that which information source should be fed to the models to enable them to make acceptable decisions. Technical analysis tools were the best candidates. Many researchers tried to verify that technical trading rules are appropriate for making investment decisions on financial assets. Different technical trading rules such as moving-average \cite{brock1992simple}, trading rage break \cite{brock1992simple}, and more complex indicators such as Japanese Ichimoku Kinkohyo \cite{deng2020profitability} were investigated and proved that the technical strategies are able to make profit. Further investigations were accomplished to verify the performance of technical strategies in different markets \cite{grobys2020technical} \cite{gunasekarage2001profitability}. 

While a lot of research is done to verify the performance of technical trading strategies in different financial markets, the essence of learning different rules for different markets became obvious. For example, the trading rules that were performing well in predicting stock price movements in the emerging markets, have less explanatory power in more developed markets \cite{bessembinder1995profitability}. Also, the appropriate indicators to capture the sell signal do not properly capture the buy signal on the same asset \cite{pramudya2020efficiency}.

The first attempts to propose a model to learn trading rules on single specific financial assets was based on genetic programming. Genetic programming was widely used to learn technical trading rules for different indices like S\&P 500 index\cite{allen1999using}, to learn appropriate trading rules to benefit from short-term price fluctuations \cite{potvin2004generating}, to learn noise-tolerant rules based on a large number of technical indicators \cite{chien2010mining}, and to learn the trading rules based on popular technical indicators like MACD \cite{mallick2008empirical}.

One of the most important weaknesses of genetic algorithms is that they are not able to evolve after task execution, while the reinforcement learning (RL) based models are able to change during the task execution. Therefore, many researches tried to combine the RL method with genetic algorithms to benefit from the advantages of both of them \cite{chen2007genetic} \cite{yang2010gnp} \cite{fischer2018reinforcement}.

Nowadays, considering the brilliant performance of RL based models in trading financial assets, the proposed models for learning stock trading rules are mainly based on the RL techniques. There are three types of RL techniques which are used in the proposed models: 1) Value-based methods in which the agent first estimates the value of each action in each state and then selects the action with highest value at each state, like Q-learning, 2) Policy-based methods in which the agent directly learns the policy function, like Policy Gradient (PG), 3) Actor-critic methods in which the actor generates an action at each time-step and the critic measures the quality of the generated action.

The following challenges should be faced in the models to learn trading rules for a specific financial asset:
\begin{enumerate}
	\item Dynamic environments:
	The financial markets are extremely dynamic and the proposed models should be able to adapt themselves quickly with the changes in the market. The proposed models should be able to continuously learn from the model and improve their parameters and performance.
	\item Models should fit on each asset in different markets:
	The proposed models should be able to learn appropriate trading rules and strategies for different assets in different markets using the asset-specific history of price data.
	\item Feature extraction:
	One of the most important parts of the proposed models to learn a good trading rule for a specific financial asset, is the feature extraction phase. The quality of the extracted features directly influences the performance of the learned trading rules.
\end{enumerate}

In this paper, we propose a deep reinforcement learning based model to generate single asset trading signals which outperforms the state-of-the-art methods. Furthermore different neural network structures are proposed for the feature extraction phase and the performance of each neural network is evaluated. In addition, the performance of DRL based models in learning asset-specific trading rules is studied.

\section{Related Work}
\label{sec:related}

The proposed approaches for single stock timing and trading strategies can be divided into two categories: 1) Knowledge-based methods in which trading strategies are designed based on mathematics or the experiences of human experts, 2) machine learning methods in which the strategies are learned from the available historical data \cite{wang2017deep}. Lee et al. \cite{lee2012causal} showed that the knowledge-based methods are not portable enough to be used as a general trading strategy in financial markets. The limitations of human reasoning may result in poor planning, hence, the system qualities of knowledge-based methods should be carefully investigated before the execution. On the other hand, since the machine-learning approaches learn the trading strategies based on the provided historical data of each asset, they can extract more profitable patterns that human traders cannot find out conveniently.

Among different methods and techniques in machine learning, genetic algorithms and reinforcement learning are those which are used to learn single stock trading rules more frequently. Allen et al. \cite{allen1999using} proposed one of the first methods based on genetic algorithms to learn trading rules for the S\&P 500 index using daily prices. Even though, the method proposed by Allen et al. \cite{allen1999using} was unable to earn consistent excess returns over the buy-and-hold strategy after considering trading costs, it was a start point for continuing research on learning more profitable trading rules by genetic algorithms. Potvin et al. \cite{potvin2004generating} proposed a model based on genetic programming to exploit the short-term fluctuations on the individual stocks which was able to make profitable trades rather than the buy-and-hold strategy. Mallick et al. \cite{mallick2008empirical} also proposed a model based on genetic programming to automatically generate trading rules on the single stocks in different market scenarios. The model proposed by Mallick et al. \cite{mallick2008empirical} was also to ensure a positive dollar return on thirty component stocks of the Dow Jones Industrial Average index. Chien et al. \cite{chien2010mining} proposed a model based on genetic algorithm which was able to create an associative classifier to classify each time step to one of sell or buy situations.

Better performances gained by employing genetic algorithms for single stock timings motivated the community to improve the quality of such models. Chen et al. \cite{chen2010model} proposed a time adapting genetic network programming model which was able to cope with the temporal behavior of asset prices. Chen et al. \cite{chen2011genetic} employed the genetic relation algorithm and considered the correlation coefficient between stock brands as the edges in a graph structure to pick up the most efficient portfolio. Michell et al. \cite{michell2020strongly} proposed a combination of the fuzzy inference system and strongly typed genetic programming to improve the efficiency of the genetic programming techniques. Michell et al. \cite{michell2020generating} also proposed a model based on strongly typed genetic programming to generate single stock trading rules focusing the fitness function on a ternary decision based on the return prediction of the corresponding stock.

One of the most important weaknesses of the genetic algorithms is that they cannot perform well in dynamic environments. One approach to improve the tolerance of such models in financial markets, which are extremely dynamic, is to combine them with different reinforcement learning techniques. Chen et al. \cite{chen2007genetic} combined the genetic network programming (GNP) with SARSA learning algorithm in order to enable the evolution-based method to change the program during the task execution. Yang et al. \cite{yang2010gnp} added some subroutines to GNP-SARSA algorithm which is able to call a corresponding subprogram during the execution. Fischer et al. \cite{fischer2018reinforcement} also proposed a GNP-SARSA based model with plural subroutines with different structure  in which each subroutine node could define its own input and output nodes.

The reinforcement learning has the following three major benefits over other machine-learning approaches which attracted the attention of the community:

\begin{enumerate}
	\item It needs no prior knowledge of the environment to learn the trading rules.
	\item It is able to continuously adapt itself to the new situations of the environment.
	\item It considers long-term benefits rather than immediate returns.
\end{enumerate}

The above-mentioned advantages of the reinforcement learning encouraged the research community to think of the RL method combined with the deep neural networks, which are able to extract rich features from the environment, as an stand alone approach to learn appropriate trading strategies. Recent research showed that the combination of deep neural networks and reinforcement learning yields an extremely powerful model to learn a good policy without any knowledge about the environment. Mnih et al. \cite{mnih2015human} described the Deep Reinforcement Learning (DRL) method of Google's DeepMind team which was able to play seven different Atari games and even defeated the human top players in three of them.

Recently, DRL models are widely used to learn a good single stock trading strategy for a given stock based on its historical data. Deng et al. \cite{deng2016deep} proposed a model based on a recurrent deep neural network trained with reinforcement learning for real-time financial signal representation in an unknown environment. Wang et al. \cite{wang2017deep} proposed a model based on deep Q-learning to build an end-to-end system for taking good positions at each trading time step. Xiong et al. \cite{xiong2018practical} employed the Deep Deterministic Policy Gradient (DDPG) technique to learn a dynamic stock trading strategy which outperformed the Dow Jones Industrial Average and the min-variance portfolio allocation. Li et al. \cite{li2019application} examined the performance of three variations of Deep Q-network including typical DQN, Double DQN, and Dueling DQN in learning single stock trading strategies for ten US stocks and concluded that the typical DQN maximizes the decisions benefits over three methods. Luo et al. \cite{luo2019novel} combined two convolutional neural networks (CNNs) as feature extractors with a DDPG model for learning trading strategies on real stock-index future data. Zarkias et al. \cite{zarkias2019deep} proposed a novel price trailing method by reformulating trading as a control problem and leaned trading strategies based on trend following for taking profitable decisions. Zhang et al. \cite{zhang2020deep} employed RL to design trading strategies for future contracts and investigated both discrete and continuous action spaces with a reward function modified by a volatility scaling. The authors showed that the RL method and the modern portfolio theory are equivalent if a linear utility function is used. Theate et al. \cite{theate2020application} used Sharpe ratio performance indicator as the reward function in his proposed model.

Research on DRL models for learning trading strategies showed that the performance of the proposed models is highly dependent on the information quality of their inputs. Therefore, some researchers tried to extract temporal dependencies of price time-series to improve their proposed models. Wu et al. \cite{wu2020adaptive} used a Gated Recurrent Unit (GRU) to extract temporal dependencies from raw financial data and technical indicators in combination with the DQN and Deterministic Policy Gradient (DPG) models to learn a trading strategy on single stocks. Suchaimanacharoen et al. \cite{suchaimanacharoen2020empowered} first predicted the future prices of a currency pair (EUR/USD) using a CNN and then fed the forecasting prices to the Policy Gradient (PG) model to learn a trading strategy in the high frequency trading domain. Lei et al. \cite{lei2020time} proposed a time-driven feature-aware jointly deep reinforcement learning model called TFJ-DRL which was able to learn feature representation from highly non-stationary and noisy environments and extract the temporal dependencies in an online manner, simultanously. 

Even though, a wide variety of time-series models are employed to provide better representations of price movements from the historical data, the problem still remains unresolved and proposed models are not capable of learning a well-qualified feature vector to consider previous price movement behavior. One of the useful financial representation techniques to demonstrate the price movement behavior in a short period of time, is the Japanese Candlestick charts. Candlestick charting is one of the oldest methods to demonstrate the price rises and falls during a period of time which were proposed first by a Japanese merchant, Munehisa Homma, to predict the changing prices of rice \cite{northcott2009complete}. Hu et al. \cite{hu2018deep} proposed a novel investment decision strategy using convolutional auto-encoder learning stock representation from candlestick charts. Thammakesorn et al. \cite{thammakesorn2019generating} proposed a model for generating stock trading strategies employing features that effectively can recognize good patterns in candlestick charts. Orquin et al. \cite{orquin2020predictive} tested the efficiency of EUR/USD pair taking only into consideration the candlestick charts of the prices. Birogul et al. \cite{birogul2020yolo} used the famous YOLO (You Only Look Once) object detector to detect the patterns in candlestick charts in order to generate Buy/Sell signals for a stock. Fengqian et al. \cite{fengqian2020adaptive} proposed a novel technique to generate trading strategies on single stocks using candlestick charts. In this study, an RL technique is proposed to learn stock timing given the current pattern detected from the candlestick charts of the price at each time step. The pattern detection in this study is accomplished by clustering similar candlestick patterns using K-means algorithm.

In this work, we investigated the performance of SARSA($\lambda$) as a traditional RL technique for learning more fitted trading rules based on candlestick chart patterns for single stocks. 
Furthermore, a neural network based on DQN model is proposed to improve the expected return of the learned trading rules. In addition, an extra layer for learning patterns from raw OHLC prices that are better than popular candlestick chart patterns is proposed and the performance of different structure for this layer is studied. In the end, an end-to-end model for learning a trading strategy for each single asset or derivative based on its historical price data is proposed.

\section{Proposed Method}
\label{sec:method}

\subsection{Trading rules based on candlestick chart patterns}

\begin{figure}
	\centering
	\includegraphics[scale=0.2]{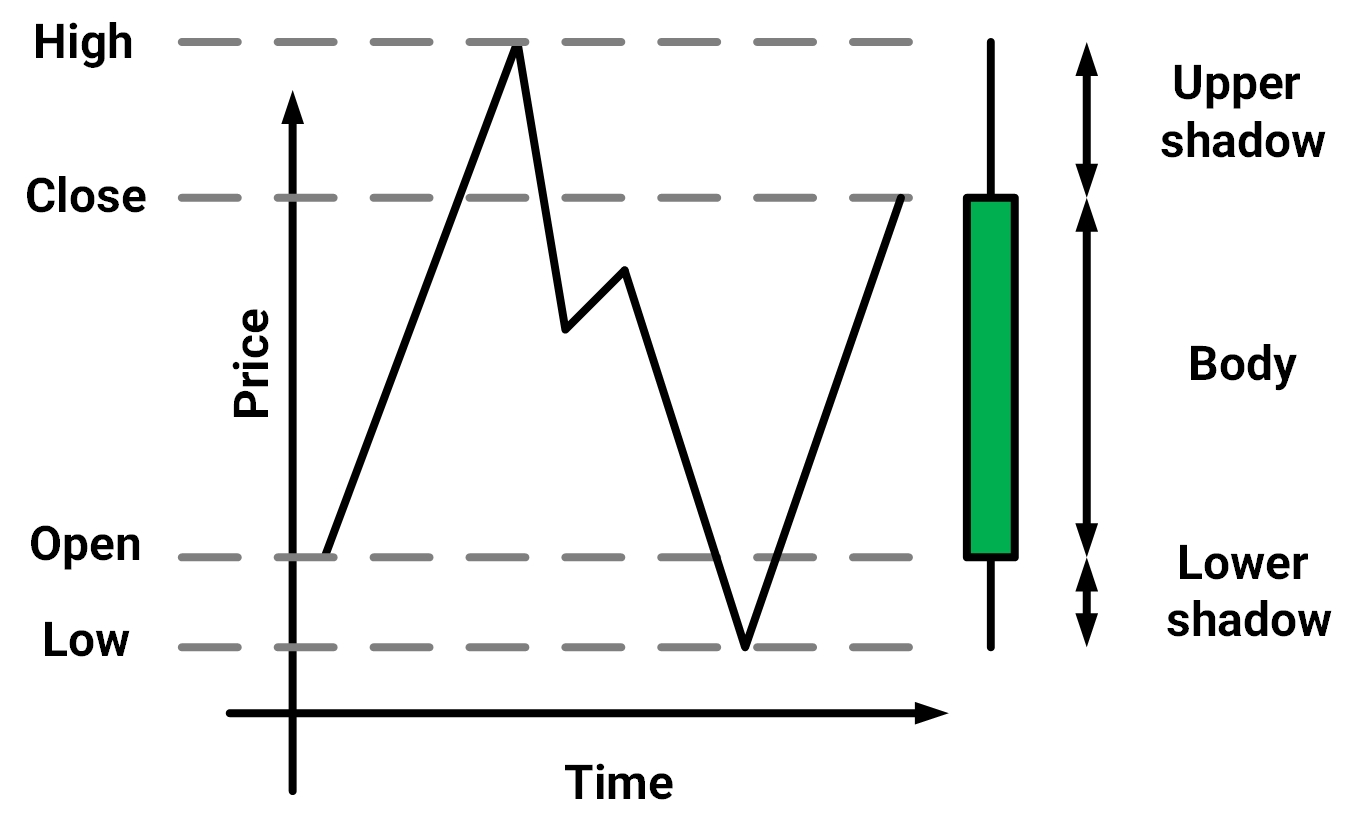}
	\caption{A candlestick representing the price behavior of an asset during a specific time window}
	\label{fig:candle}
\end{figure}

A candlestick chart is used to demonstrate the price behavior of a financial asset during a certain time window. A candlestick which is shown in figure \ref{fig:candle} is consisted of a line demonstrating the highest and the lowest prices of the asset, and a body demonstrating the first (open) and the last (close) prices during a specific time period. Typically, if the closing price is higher than the opening price, the candlestick is colored in green or white (denoting a bullish pattern) and otherwise it is colored in red or black (denoting a bearish pattern) to show the direction of the price changes. In this work, Each candlestick is a vector $\theta$ showing the open, high, low and close prices. Equation \eqref{eq:candle-vector} shows the candlestick vector at time step t, consisting of OHLC price. For simplicity, the close price of a candle stick is denoted by $P$.

\begin{equation}
c_t = (p_{open}, p_{high}, p_{low}, p_{close})
\label{eq:candle-vector}
\end{equation}

In some cases, the candlesticks form popular patterns showing a special emotional situation of the traders that can be used for analysis or trading purposes. Based on these popular patterns, a list of trading rules are extracted empirically to trade single assets in financial markets. A list of such trading rules is reported in tables explained in \eqref{sec:app1} \cite{keller2006candlestick}. 

Note that, in almost all of the trading rules based on candlestick charts another parameter denoting the current market trend is used. In this work, the market trend is detected using equation \eqref{eq:trend} in which the $\mu_w$ denotes the $w$ day moving average which is computed by equation \eqref{eq:MA}, and $v$ is a window size for the number of candles before time $t$ to be considered in the calculation of the trend. 

\begin{equation}
MT = \begin{cases}
\text{uptrend}
&\begin{split}
\quad\text{if } &\forall i \in \{0 \dots v\} \\
&\mu_w(t - i - 1) \leq \mu_w(t - i)\\
\end{split}\\
\\
\text{downtrend} 
&\begin{split}
\quad\text{if } &\forall i \in \{0 \dots v\} \\
&\mu_w(t - i - 1) \geq \mu_w(t - i)\\
\end{split}\\
\\
\text{Side} &\quad\text{otherwise}\\
\end{cases}
\label{eq:trend}
\end{equation}


\begin{equation}
	\mu_w = \frac{P_{t-w} + P_{t-w+1} + \dots + P_t}{w}
	\label{eq:MA}
\end{equation}

A signaling function $\psi_t$ is defined in equation \eqref{eq:signals} which generates stock signals at each time step based on the occurred candlestick pattern at time $t$.

\subsection{Trading rules learned by SARSA ($\lambda$) algorithm}

The first learning algorithm which is used to learn fitted trading rules on single assets in this work, is modeled by the SARSA($\lambda$) algorithm. The SARSA algorithm was proposed by Rummery et al. \cite{rummery1994line} in 1994 as a new modification of Temporal Difference (TD) algorithm. In this work, SARSA($\lambda$) is used which is an off-policy version of the single-step SARSA. In this algorithm the accumulated reward function is formed as in equation \eqref{eq:SARSAG} \cite{sutton1998introduction} in which $T$ denotes the final time step in an episode, $R_t$ denotes the immediate reward at time step $t$, $\gamma$ denotes the discount rate, $\hat{q}$ denotes the estimate of the action-value function, $S_t$ denotes the state vector at time step $t$, $A_t$ denotes the selected action at time $t$, and $w_t$ denotes the trainable parameters of the model at time step $t$.

\begin{equation}
\begin{split}
	G_{t:t+n} &= R_{t+1} + \cdots + \gamma^{n-1}R_{t+n}+\gamma^n\hat{q}(S_{t+n}, A_{t+n}, w_{t+n-1})\\
	G_{t:t+n} &= G_t \>\>\> if \>\> t+n > T\\
\end{split}
\label{eq:SARSAG}
\end{equation}

The parameter's update rule in the SARSA($\lambda$) algorithm follows the update rule structure in the temporal difference learning methods which is displayed in equation \eqref{eq:update}, in which $\eta$ is the algorithm learning rate, $\delta_t$ is the temporal difference error for action-value estimation computed by equation \eqref{eq:delta}, and $z_t$ demonstrates the action-value eligibility trace vector computed by equation \eqref{eq:zt}.

\begin{equation}
	w_{t+1} = w_t - \eta \delta_t z_t
	\label{eq:update}
\end{equation}

\begin{equation}
	\delta_t = R_{t+1} + \gamma \hat{q}(S_{t+1}, A_{t+1}, w_{t})+ \hat{q}(S_{t}, A_{t}, w_{t})
	\label{eq:delta}
\end{equation}

\begin{equation}
\begin{split}
	z_{-1} &= 0 \\
	z_t &= \gamma \lambda z_{t-1} + \Delta \hat{q}(S_{t}, A_{t}, w_{t})
\end{split}
\label{eq:zt}
\end{equation}

The immediate reward $R_t$ here is defined as in equation \eqref{eq:Rt} in which TC is the transaction cost, $P_2$ is the end price (say after n steps), and $P_1$ is the current price.
\begin{equation}
	R_t = \begin{cases}
	((1 - TC)^2 \times \frac{P_2}{P_1} - 1) \times 100 \quad\text{if action = buy}\\
	((1 - TC)^2 \times \frac{P_1}{P_2} - 1) \times 100 \quad\text{if action = sell}\\
	\end{cases}
	\label{eq:Rt}
\end{equation}

The SARSA($\lambda$) agent, takes a state vector $s_t \in S$ and an immediate reward $R_t$ from the environment at each time step $t$, and produces an action $a_{t+1} \in A$ for the next time step. The state space here consists of candlestick vectors with specific types. These types are explained in section \ref{sec:input-types}.
The action space $A$ denotes a set of three discrete actions $A = \{'sell', 'buy', 'idle'\}$.
Algorithm \ref{alg:euclid} briefly explains the \textit{Value Iteration} used to train. 

\vspace{0.5cm}
\begin{minipage}{\linewidth}
	\begin{algorithm}[H]
		\caption{The SARSA($\lambda$) algorithm used in this project}\label{alg:euclid}
		\begin{algorithmic}[1]
				\State define n as in n-step SARSA
				\State i = 0
				\While {i < size(data) - n}
				\State current-state = data[i]
				\State action = $\epsilon$-Greedy(current-state) \textit{if} current-state $\neq$ 'idle' \textit{else} 'None'
				\State next-state = data[i + n]
				\State next-action = Greedy(next-state) \textit{if} next-state $\neq$ 'idle' \textit{else} 'None'
				\State $\text{Q[current-state][action]} = (1 - \alpha) \times  \text{Q[current-state][action]} - \alpha \times 
				(Reward_n(i) + \gamma^n \times \text{Q[next-state][next-action]})$ 
				\EndWhile	
		\end{algorithmic}
	\end{algorithm}
\end{minipage}
\vspace{0.5cm}

\subsection{Trading rules learned by Deep Q-network agent}
Even though the SARSA($\lambda$) model is able to learn better signaling rules when a certain popular candlestick pattern has occurred on a single specific financial asset, it is not possible to learn more general rules for cases that an unpopular candlestick pattern has occurred or more than one pattern appears in the input. To solve this problem, it is required to employ deep neural networks which are powerful in extracting rich feature vectors and take complex trading decisions.

\begin{figure*}[htb]
	\centering
	\includegraphics[scale=0.15]{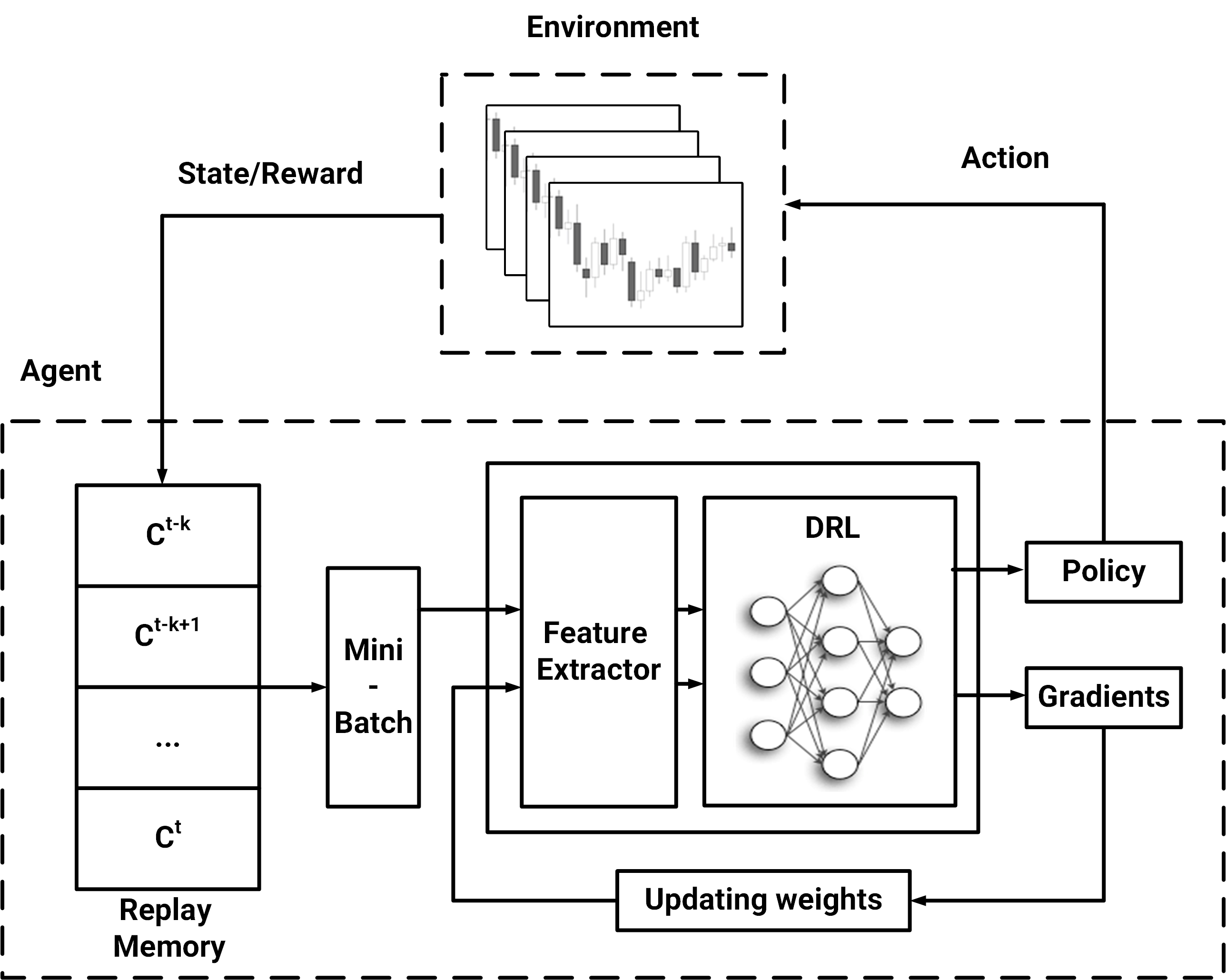}
	\caption{In this architecture, at each time step, the state is given by the environment, and the agent takes action according to the state and receives the reward and next state. The quadruples $(CurrentState, Action, Reward, NextState)$ are saved in the replay memory. For the optimization part in each iteration, after the aforementioned step, a batch of quadruples $(CurrentState, Action, Reward, NextState)$ is selected for the training. The replay memory has a specific capacity and after it is filled, a random quadruple is substituted with a new quadruple}
	\label{fig:arch}
\end{figure*}

In order to address this issue, a model based on the architecture of the DQN\cite{mnih2015human} is proposed which can extract rich feature vectors from the input time-series and learn good trading signals for the corresponding asset. The architecture of the proposed model is displayed in Figure \ref{fig:arch}.

The proposed model consists of two modules: feature extraction, and decision making. In this architecture, the input time-series of raw OHLC prices is passed to the first module in order to extract the candlestick pattern and form a suitable feature vector based on the corresponding candlestick. The extracted feature vector is then passed to the decision making module which is designed based on the DQN network. The output of this module is the action that agent should take at the next time-step. It is worth noting that the next state vector and the corresponding immediate reward are passed to the decision making module directly from the environment. The architectures of these modules are discussed in detail in the following sections.

\subsubsection{Decision making}
In order to strengthen the proposed model to learn more powerful rules which can generate appropriate signals when an unpopular candlestick pattern is occurred, it is necessary to use a deep neural network to estimate the state action-value function (Q function) instead of using a lookup table-based approach. The decision making module in the method proposed here is structured based on the deep reinforcement learning frameworks proposed by Moody et al. \cite{moody1998performance} and Mnih et al. \cite{mnih2015human}. Suppose the target Q function is parameterized with a set of network weights $\Theta$. According to the Bellman equation, it is possible to train the corresponding network by equation \eqref{eq:VanillaDQN}, in which the $L_\Theta$ denotes the network's loss function, $Q_\Theta(S, A)$ denotes the target Q function, and $\hat{Q_\Theta}(S, A)$ denotes the estimated Q function at each time step.

\begin{equation}
	L(\Theta) = E((Q_\Theta(S, A) - \hat{Q_\Theta}(S, A))^2)
	\label{eq:VanillaDQN}
\end{equation}

The $\hat{Q_\Theta}(S, A)$ function can be estimated using the Bellman equation as in equation \eqref{eq:bellman}.

\begin{equation}
	\hat{Q_\Theta}(S_t, A_t) = R_t + \gamma \> argmax_{\hat{A}} Q_{\Theta}(S_{t+1}, A_{t+1})
	\label{eq:bellman}
\end{equation}

In order to stabilize the training process, a frozen copy of the network is created at the end of each training iteration in which the weights are not being changed during each training iteration. The value of the $Q_{\Theta}(S_{t+1}, A_{t+1})$ in equation \eqref{eq:bellman} then, is estimated with applying this frozen network to the selected action at the current time step.

The neural network which is used to estimate the Q function, consists of three fully connected layers. The first layer, transforms the input space to a 128-dimensional latent space, the second layer transforms the 128-dimensional space to a second 256- dimensional latent space, and the last layer converts the seconds space to a 3-dimensional action space. In order to stabilizing the network and avoiding overfitting, a batch normalization layer is used between the layers. In addition, a \textit{Softmax} layer is used to generate a probability distribution over the action space at the last layer of the network.

\subsubsection{Feature extraction}
The feature extraction phase of the proposed model is accomplished by a neural network which is trained jointly with the decision making module. The back propagated error in decision making module is directly used as the error at the last layer of the feature extraction module. 

Different network structures including MLP, CNN, and GRU architectures are used to find the best model to extract rich features for signal generation. Since the input here is not a long sequence of asset prices (in most cases the candlestick pattern of the current time step and in some cases the candlestick patterns appeared at the last three time step are used as input), simpler networks reached better results. 

The output of the feature extractor is concatenated with the market trend extracted by equation \eqref{eq:MA} and the resulting vector is passed to the decision making module to generate stock trading signals.

\subsubsection{Input vectors}
To study the effect of different input representations on the performance of the model, three types of inputs are provided and fed separately to the proposed model. In the first method, a list of popular candlestick patterns are provided and at each time step a binary vector specifying the appeared candlestick pattern is computed. In the second form, the percentage of the upper shadow, the body, and the lower shadow of the candlestick are computed as displayed in equation \eqref{eq:candles} and passed to the model. In the third method, the raw values of the OHLC prices are fed to the model. In the third method, the feature extraction module is responsible for learning a good representation of the input.

\begin{equation}
\begin{split}
&upper = \frac{p_h - max(p_c, p_o)}{p_h - p_l} \\
&lower = \frac{min(p_c, p_o) - p_l}{p_h - p_l} \\
&body = \frac{|p_c - p_o|}{p_h - p_l} \\
\end{split}
\label{eq:candles}
\end{equation}

\subsubsection{Training algorithm}
The process of training the model is displayed in algorithm \ref{alg:q-learning}. 

\vspace{0.5cm}
\begin{minipage}{\linewidth}
	\begin{algorithm}[H]
		\caption{Deep Q-Learning Algorithm used for training}\label{alg:q-learning}
		\begin{algorithmic}[1]
				\State Initialize replay memory D to capacity N
				\State Initialize action-value function Q with random weights $\theta$
				\State Initialize target action-value function $\hat{Q}$ with weights $\theta^- = \theta$
				
				\For{episode from 1 to M}
				\State Initialize sequence $s_1$ and preprocessed sequence $\phi_1 = \phi(s1)$
				\For{t from 1 to T}
				\State With probability $\epsilon$ select a random action $a_t$
				\State Otherwise select $a_t = argmax_a Q(\phi(s_t), a; \theta)$
				\State Execute action $a_t$ and observe reward $r_t$ and state $s_{t+1}$
				\State Set $s_{t+1} = s_t, a_t$ and preprocess $\phi_{t+1} = \phi(s_{t+1})$
				\State Store transition ($\phi_t, a_t, r_t, \phi_{t+1}$)
				\State Sample random mini-batch of transitions ($\phi_j, a_j, r_j, \phi_{j+1}$) from D
				
				\State \begin{align*}
				Set \, y_j =
				\begin{cases}
				r_j \quad\text{if episode terminates at step j + 1}&\\
				r_j + \gamma \max_{a'} \hat{Q}(\phi_{j+1}, a'; \theta^-) \quad\text{otherwise}& \\
				\end{cases}
				\end{align*}
				
				\State Perform a gradient descent step on $(y_j - Q(\phi_j, a_j; \theta))^2$
				with respect to the network parameters $\theta$
				\State Every C steps reset $\hat{Q} = Q$
				\EndFor
				\EndFor
		\end{algorithmic}
	\end{algorithm}
\end{minipage}
\vspace{0.5cm}

\section{Experimental Results}
\label{sec:experiment}
\subsection{Datasets}

\begin{figure*}
	\centering
	\begin{subfigure}{\textwidth}
		\includegraphics[width=\linewidth, height=0.20\textheight]{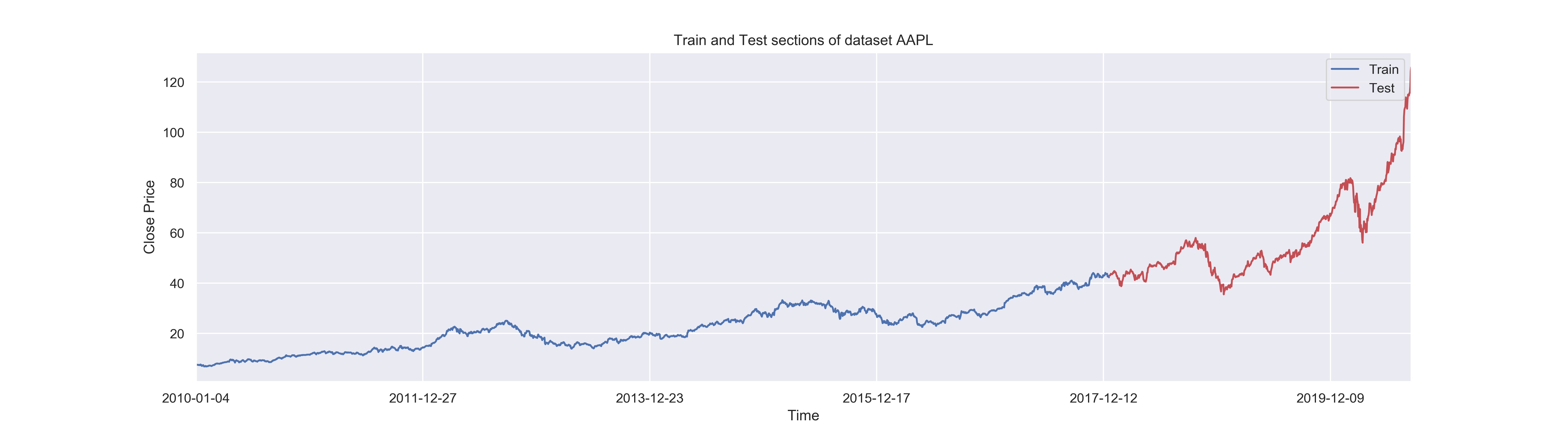}
		\caption{Price history of AAPL stock used to train and test the model.}
	\end{subfigure}
	\\
	\begin{subfigure}{\textwidth}
		\includegraphics[width=\linewidth, height=0.20\textheight]{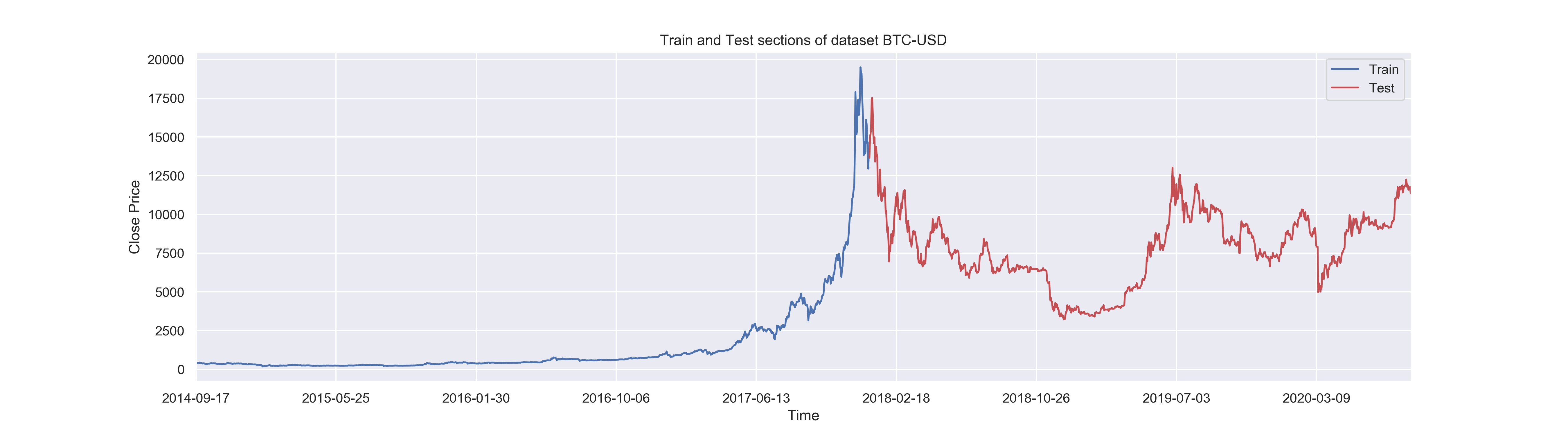}
		\caption{Price history of BTC/USD pair used to train and test the model.}
	\end{subfigure}
	\\
	\begin{subfigure}{\textwidth}
		\includegraphics[width=\linewidth, height=0.20\textheight]{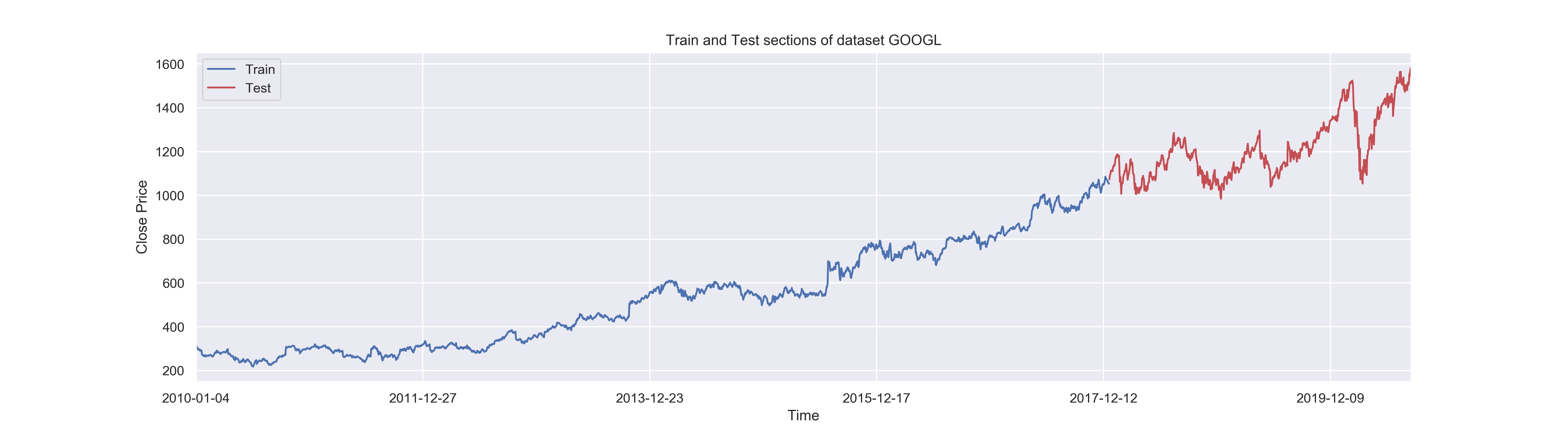}
		\caption{Price history of GOOGL stock used to train and test the model.}
	\end{subfigure}
	\begin{subfigure}{\textwidth}
		\includegraphics[width=\linewidth, height=0.20\textheight]{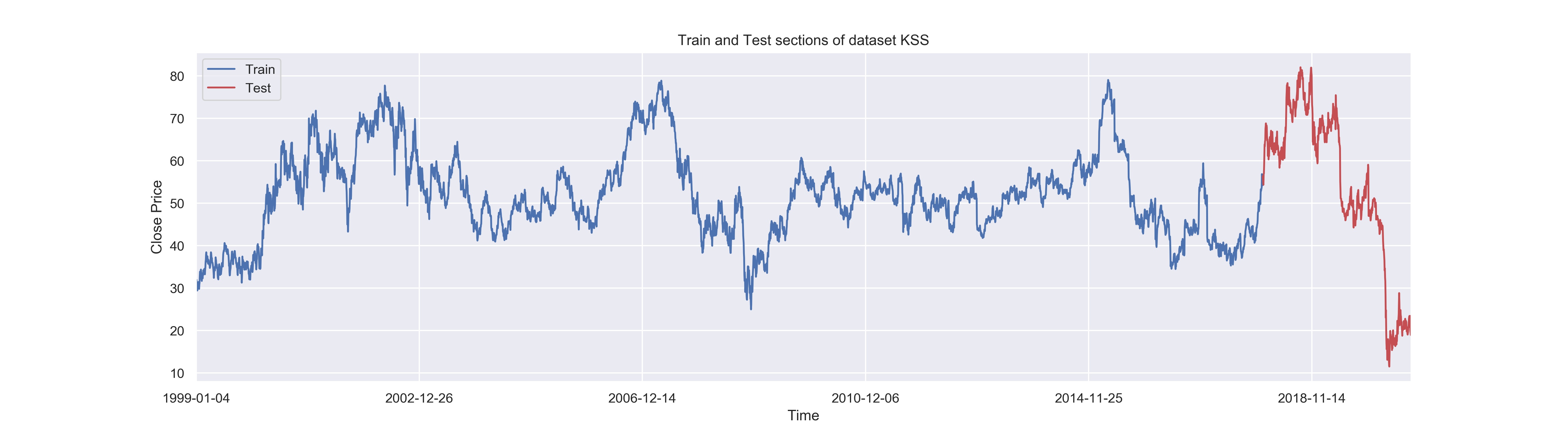}
		\caption{Price history of KSS stock used to train and test the model.}
	\end{subfigure}

	\caption{Price histories used to test the models. The blue sections are used for training and the red parts are used as testing sets.}
	\label{fig:dataset}
\end{figure*}

The proposed methods are tested on the real-world financial data including the stocks, currency pairs, and crypto-currencies data. For the stock data we select the historical data of AAPL, GOOGL, and KSS. 
For crypto-currencies data, we chose BTC/USD which is the price of Bitcoin. All the data used in this paper are available at \textit{Yahoo Finance} and \textit{Google Finance}.
All of the candlesticks are created in daily time window according to table \ref{tbl:data}, the price history of each asset is divided into two parts, which is displayed in figure \ref{fig:dataset}.

\begin{table}[htb]
	\centering
	\caption{Data used along with train-test split dates}
	\begin{tabular}{|c||c|c|c|}
		\hline
		Data & Begin Date & Split Point & End Date \\
		\hline\hline
		GOOGL & 2010/01/01 & 2018/01/01 & 2020/08/24\\ 
		AAPL & 2010/01/01 & 2018/01/01  & 2020/08/24\\  
		KSS & 1999/01/01 & 2018/01/01 & 2020/08/24\\  
		BTC-USD & 2014/09/17 & 2018/01/01 & 2020/08/26\\
		\hline    
	\end{tabular}
	\label{tbl:data}	
\end{table}

Considering the available price histories from figure \ref{fig:dataset}, extremely different assets with highly variant price movement behaviors appear in the testing data. The history used to train is illustrated in blue and the rest of the data used to test the model is illustrated in red. The AAPL and the GOOGL data have ascending trends in both training and testing parts. The BTC/USD dataset has an ascending training history while the testing part is started with an extremely descending behavior which is turned to a side trend after that. The KSS stock has a side trend in training while in the testing set a descending trend has appeared. The interesting point about the BTC/USD pair is that the model is given an ascending trend during the training, while it is tested on a descending trend while testing. Hence, the performance of the model on the KSS stock shows the generalizability of learned trading rules. Different price movement behaviors in training and testing are used to evaluate the ability of the model to learn good trading rules in even previously unknown market situations.

\subsection{Evaluation metrics}
The proposed model is evaluated with respect to two types of evaluation metrics: 1) metrics related to the profitability of the learned training rules, 2) metrics related to the implied risk of investment based on the learned trading rules.
More details about the evaluation metrics are as follows.
\begin{enumerate}[i)]
	\item Daily returns: 
	The sequence of daily returns are computed by equation \eqref{eq:ARLR}, and their average value over the testing and training period is reported. In these equations $\Omega_t$ denotes the total value of the portfolio at time step $t$.
	\begin{equation}
	\begin{split}
		&AR_t = \frac{\Omega_t - \Omega_{t-1} }{ \Omega_{t-1} }\\
	\end{split}
	\label{eq:ARLR}
	\end{equation}
	
	\item Total return:
	The ratio of the capital growth during testing and training time. The total return is computed by equation \eqref{eq:TT} in which $\Omega_{0}$ denotes the initial investment and $\Omega_{T}$ denotes the value of the portfolio at the end of the period.
	\begin{equation}
		TT = \frac{\Omega_T - \Omega_{0} }{ \Omega_{0} }
		\label{eq:TT}
	\end{equation}
	
	\item Value at risk: 
	The value at risk ($VaR$) is a metric to measure the quality level of a financial risk within a portfolio during a specific period of time. $VaR$ typically is measured with a confidence ratio $\alpha \in (-1, 1)$ and measures the probability of gaining a return less than $\alpha$ in the corresponding time period. The higher the value of the $VaR_\alpha$ with a fixed value of $\alpha$, the higher the level of financial risk of the portfolio. There exists two main approaches to compute $VaR_\alpha$: 1) using the closed form which assumes the probability distribution of the daily returns of the portfolio follows a Normal standard distribution, 2) using the historical estimation method which is a non-parametric method and assumes no prior knowledge about the portfolio's daily returns. In this paper, we used the closed form method. To calculate $VaR_{\alpha}$, 
%
we used Monte Carlo simulation by developing a model for future stock price returns and running multiple hypothetical trials through the model. The mean $\mu$ and standard deviation $\sigma$ of the returns are calculated, then 1000 simulations run to generate random outputs with a normal distribution $N(\mu, \sigma)$. Then the $\alpha$ percent lowest value of the outputs is selected and reported as $VaR_\alpha$.

	\item Daily returns volatility:
	The volatility of the daily returns tells us about the financial risk level of the trading rules. The volatility is estimated using the standard deviation of the time series of the daily returns of the portfolio and is computed by equation equation \eqref{eq:volume}.
	
	\begin{equation}
		\sigma_p = \sqrt{\frac{1}{T-1} \Sigma_{i=1}^T (AR_i - \mu(AR_{1:T}))^2}
		\label{eq:volume}
	\end{equation}
	
	\item Sharpe ratio:
	The Sharpe ratio (SR) was proposed first by Sharpe et al. \cite{sharpe1994sharpe} to measure the reward-to-variability ratio of the mutual funds. This metric displays the average return earned in excess of the risk-free rate per unit total risk and is computed here by equation \eqref{eq:SR} in which $R_f$ is the return of the \textit{risk-free} asset, and $E\{R_p\}$ is the expected value of the portfolio value. Here we assumed that $R_f=0$.
	
	\begin{equation}
		SR = \frac{E\{R_p\} - R_f}{\sigma_p}
		\label{eq:SR}
	\end{equation}
	
	
	\item Profit curve:
	The profit curve is a qualitative metric which reflects the profits gained by the model at each time step. In this paper, the profit curves of different models are drawn within a single chart in a specific period of time to compare the performance of these models with respect to their profitability during the investment period.
	
	\item Decision curve:
	In this curve, the trading signals to trade each asset is demonstrated over the raw price curve of that asset. This chart gives insight about the quality of decision making power of each model on each financial asset.
	
\end{enumerate}

\subsection{Baseline models}
Whenever possible, the proposed models in this paper are compared with the state-of-the-art models of learning single asset trading rules. Since the implementations of the most of these models are not accessible, comparison with each baseline model is accomplished just in cases that the model performance metrics are reported in the corresponding paper. The list of the used baseline models is as follows.

\begin{enumerate}[i)]
	\item \textbf{Buy and Hold (B\&H)}:
	The B\&H is one of the most widely used benchmark strategies to compare the performance of the proposed model with. In this strategy, the investor selects an asset and buys it at the first time step of the investment and holds it to the end of the period regardless of its price fluctuations.
	\item \textbf{GDQN}:
	Proposed by Wu et al. \cite{wu2020adaptive}, using the concatenation of the technical indicators and raw OHLC price data of last 9 time steps as the input, a two-layered stacked structure of GRUs as the feature extractor and the DQN as the decision making module.
	\item \textbf{DQT}:
	Proposed by Wang et al. \cite{wang2017deep}, implementing online Q-learning algorithm to maximize the long-term profit of the investment using the learned rules on a single financial asset. The reward function here is formed computing the accumulated wealth over the last $n$ days.
	\item \textbf{DDPG}:
	Proposed by Xiong et al. \cite{xiong2018practical} using Deep Deterministic Policy Gradient(DDPG) as the Deep Reinforcement Leaning approach to obtain an adaptive trading strategy. Then the performance of the model is evaluated and compared with Dow Jones Industrial Average and the traditional min-variance portfolio allocation strategy.
\end{enumerate}

\subsection{Details on training the models}
All of the models are trained using the Adam optimizer. The mini-batch training is also conducted using a batch size of 10, and the replay memory size is set to 20. The only regularization used in the experiments is the \textit{Batch Normalization}. The transaction cost is set to zero during the training process; however, it may be non-zero during the evaluation.

\subsection{General evaluations}

\subsubsection{Feature extractors}
\begin{figure*}
	\centering
	\begin{subfigure}{\textwidth}
		\includegraphics[width=\linewidth, height=0.20\textheight]{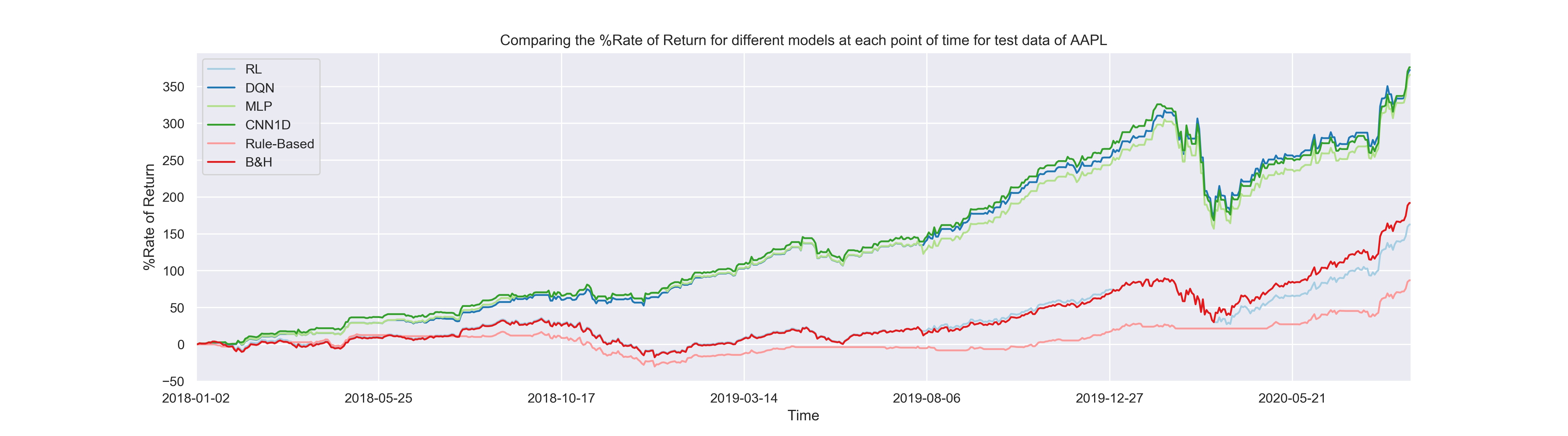}
		\caption{Performance of different models on AAPL.}
	\end{subfigure}
\\
	\begin{subfigure}{\textwidth}
		\includegraphics[width=\linewidth, height=0.20\textheight]{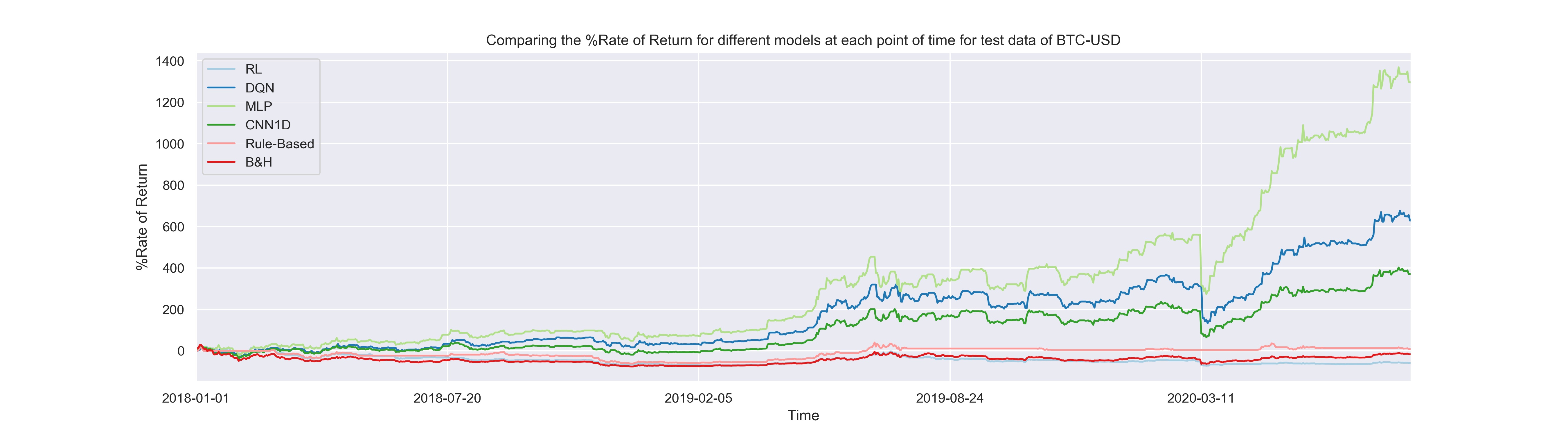}
		\caption{Performance of different models on BTC/USD pair.}
	\end{subfigure}
\\
	\begin{subfigure}{\textwidth}
		\includegraphics[width=\linewidth, height=0.20\textheight]{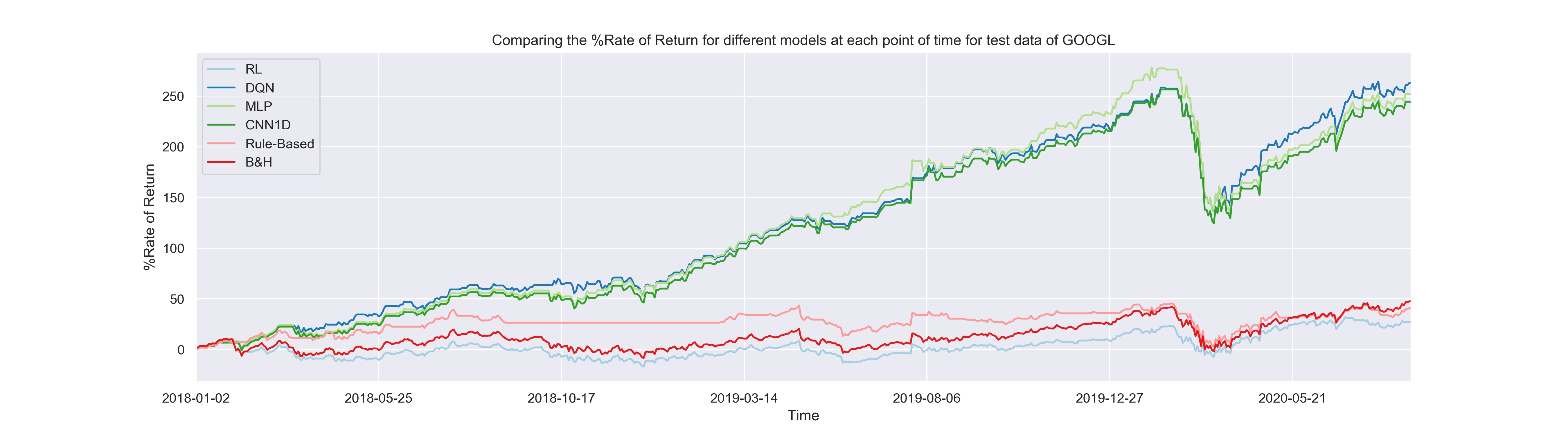}
		\caption{Performance of different models on GOOGL.}
	\end{subfigure}
\\
	\begin{subfigure}{\textwidth}
		\includegraphics[width=\linewidth, height=0.20\textheight]{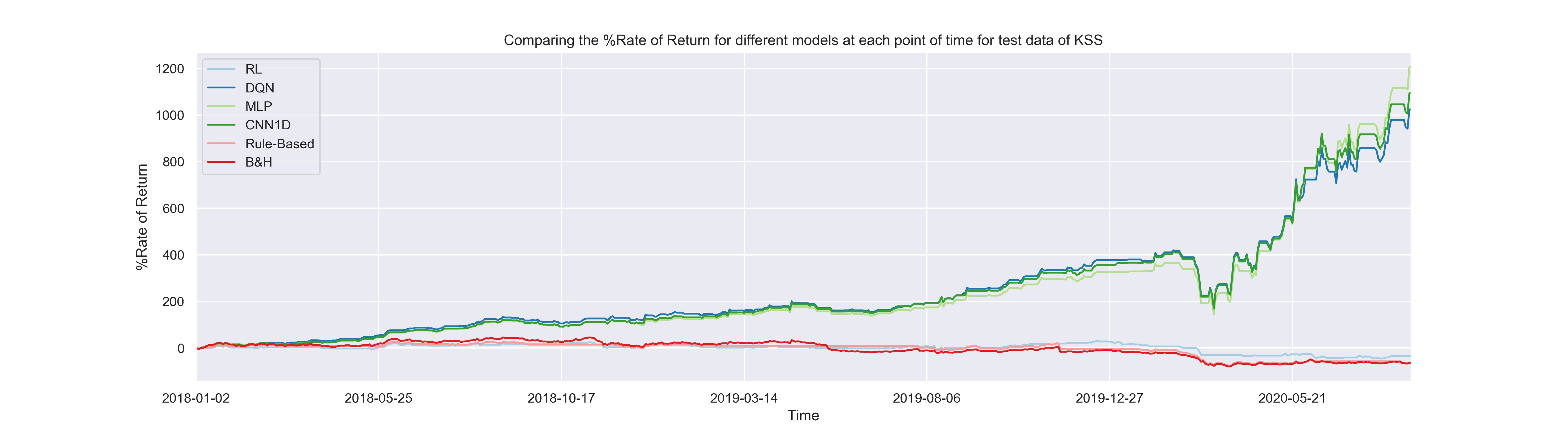}
		\caption{Performance of different models on KSS.}
	\end{subfigure}

	\caption{Profit curves of different models on each financial asset without considering different input types.}
	\label{fig:PC}
\end{figure*}

Figure \ref{fig:PC} illustrates the profit curve of different models including the experimental trading rules based on candlestick chart patterns (Rule-Based), the benchmark Buy and Hold (B\&H) strategy, the typical SARSA($\lambda$) with the optimal $\lambda = 10$ (RL), the proposed DRL method without any feature extraction layer (DQN), the DRL model with a two-layered fully connected module as the feature extractor (MLP), the DRL model with a 1-dimension convolutional layer as feature extractor (CNN1D), and the DRL model with a 2-dimension convolutional layer as feature extractor (CNN2D). Considering the experimental results, following conclusions are drawn.

\begin{enumerate}[i)]
	\item Almost in all of the four datasets, the performance of the SARSA agent, rule-based agent, and the B\&H strategy are similar while all of the DRL models, which are trained to learn trading rules specific to each of the datasets, performed much better than them. This means that learning \textbf{asset-specific} trading rules and strategies makes more profit than seeking a model to learn general trading rules which are profitable on different assets. 
	\item Between the RL-based models, models based on deep neural networks show a better performance than the SARSA model. The main difference between these models is that, the SARSA model tries to generalize the parameters of the available candlestick pattern based trading rules while the deep RL models are able to extract different new forms of rules. Furthermore, the extracted trading rules with deep RL models are more profitable than the experimental rules.
\end{enumerate}

The performance of different models with respect to the introduced metrics are reported and compared with the state-of-the-art models in learning single stock trading rules in table \ref{tbl:AAPL} on AAPL stock, table \ref{tbl:BTCUSD} on BTC/USD pair, table \ref{tbl:GOOGL} on GOOGL stock, and figure table{tbl:KSS} on KSS stock. According to the results reported in these tables, the model with the MLP feature extractor reached better total return during the testing period. One of the important points in the reported results is that the rule-based agent achieved the lowest financial risk level among the models. It is obvious that the other models learned more risk-taking trading rules because their reward function is designed based on a risk-neutral investor in which no risk-related term is appeared. The observation verifies that the form of the reward function directly impacts the behavior of the learned trading rules. If the weight of the risk-related term in the financial function is high, the learned trading rules will be appropriate for risk-averse investors.

The other important point in the results reported in tables \ref{tbl:BTCUSD}, \ref{tbl:GOOGL}, \ref{tbl:AAPL}, and \ref{tbl:KSS} is that in cases that the training set is similar to the testing set, the one dimensional convolutional feature extractor performs tightly similar to the model with MLP feature extractor, while in other datasets the MLP model outperforms the CNN1D. Also, according to table \ref{tbl:GOOGL} which illustrates the performance of models on the GOOGL with side ascending trend, the GRU model performs better than, MLP model with respect to the daily average metric. The important result here is that the feature extraction module can evidently affect the model performance and the feature extraction should be one of the most important focuses of the future research in this area.

\begin{table*}[htb]
	\centering
	\caption{Performance of different models on BTC/USD.}
	\label{tbl:BTCUSD}
	\begin{adjustbox}{width=\textwidth}
		\begin{tabular}{@{} |l*{10}{|l}| @{}}
			\hline
			Agent & \rotatebox[origin=c]{90}{Arithmetic Return} & \rotatebox[origin=c]{90}{Average Daily Return} & \rotatebox[origin=c]{90}{Daily Return Variance} & \rotatebox[origin=c]{90}{Time Weighted Return} & \rotatebox[origin=c]{90}{Total Return}  & \rotatebox[origin=c]{90}{Sharpe Ratio} & \rotatebox[origin=c]{90}{Value At Risk} & \rotatebox[origin=c]{90}{Volatility} & \rotatebox[origin=c]{90}{Initial Investment} & \rotatebox[origin=c]{90}{Final Portfolio Value}\\
			\hline
			
			Rule-Based & 45.078 & 0.04 & \textbf{7.06} & 0.000 & 7 \% & 0.016 & \textbf{-4.330} & \textbf{82} & 1000.000 & 1076 \\
			\hline
			B\&H & 60 & 0.06 & 15.78 & -0.000 & -16 \% & 0.016 & -6.480 & 123 & 1000.000 & 830 \\
			\hline
			RL      & -40 & -0.04 & 10.73 & -0.001 & -65 \% & -0.017 & -5.438 & 101 & 1000.000 & 340 \\
			\hline
			\hline
			DQN-pattern & 57 & 0.06 & 15.31 & -0.000 & -16 \% & 0.015 & -6.384 & 121 & 1000.000 & 830 \\
			\hline
			DQN-vanilla & 261 & 0.27 & 12.63 & 0.002 & 628 \% & 0.076 & -5.583 & 110 & 1000.000 & 7287 \\
			\hline
			DQN-candle-rep &‌50 & 0.05 & 15.68 & -0.000 & -24 \% & 0.013 & -6.470 & 123 & 1000.000 & 757 \\
			\hline
			DQN-windowed & ‌221 & 0.22 & 12.83 & 0.002 & 384 \% & 0.064 & -5.671 & 111 & 1000.000 & 4843 \\
			\hline
			\hline
			MLP-pattern &‌50 & 0.05 & 15.68 & -0.000 & -24 \% & 0.013 & -6.470 & 123 & 1000.000 & 757 \\
			\hline
			MLP-vanilla &\textbf{‌323} & \textbf{0.33} & 12 & \textbf{0.003} & \textbf{1295} \% & \textbf{0.097} & -5.372 & 107 & 1000.000 & \textbf{13959} \\
			\hline
			MLP-candle-rep & 50 & 0.05 & 15.68 & -0.000 & -24 \% & 0.013 & -6.470 & 123 & 1000.000 & 757 \\
			\hline
			MLP-windowed &‌221 & 0.22 & 12.62 & 0.002 & 389 \% & 0.065 & -5.622 & 110 & 1000.000 & 4892 \\
			\hline
			\hline
			CNN1D &‌219 & 0.22 & 13 & 0.002 & 370 \% & 0.063 & -5.716 & 112 & 1000.000 & 4702 \\
			\hline
			CNN2D &‌212 & 0.22 & 13.230 & 0.002 & 334 \% & 0.061 & -5.770 & 113 & 1000.000 & 4348 \\
			\hline
			\hline
			GRU & 165 & 0.17 & 13.356 & 0.001 & 168 \% & 0.047 & -5.847 & 113 & 1000.000 & 2686 \\	
			\hline
		\end{tabular}
	\end{adjustbox} 
\end{table*}

\begin{table*}[htb]
	\centering
	\caption{Performance of different models on GOOGL.}
	\label{tbl:GOOGL}
	\begin{adjustbox}{width=\textwidth}
		\begin{tabular}{@{} |l*{10}{|l}| @{}}
			\hline
			Agent & \rotatebox[origin=c]{90}{Arithmetic Return} & \rotatebox[origin=c]{90}{Average Daily Return} & \rotatebox[origin=c]{90}{Daily Return Variance} & \rotatebox[origin=c]{90}{Time Weighted Return} & \rotatebox[origin=c]{90}{Total Return}  & \rotatebox[origin=c]{90}{Sharpe Ratio} & \rotatebox[origin=c]{90}{Value At Risk} & \rotatebox[origin=c]{90}{Volatility} & \rotatebox[origin=c]{90}{Initial Investment} & \rotatebox[origin=c]{90}{Final Portfolio Value}\\
			\hline
			
			Rule-Based & 46 & 0.07 & 2.51 & 0.001 & 40 \% & 0.040 & -2.54 & 40.9 & 1000 & 1408 \\
			\hline
			B\&H & 51 & 0.07 & 3.77 & 0.001 & 47 \% & 0.040 & -3.12 & 50.1 & 1000 & 1475 \\
			\hline
			RL      & 39 & 0.06 & 3.58 & 0.000 & 27 \% & 0.028 & -3.05 & 48.8 & 1000 & 1270 \\
			\hline
			\hline
			DQN-pattern & 50 & 0.07 & 3.77 & 0.001 & 46 \% & 0.039 & -3.12 & 50.1 & 1000 & 1462 \\
			\hline
			DQN-vanilla & 137 & 0.20 & 2.59 & 0.002 & 263 \% & 0.128 & -2.44 & 41.5 & 1000 & 3631 \\
			\hline
			DQN-candle-rep & 49 & 0.07 & 3.77 & 0.001 & 45 \% & 0.039 & -3.12 & 50.1 & 1000 & 1452 \\
			\hline
			DQN-windowed & 152 & 0.22 & 2.45 & 0.002 & 324 \% & 0.147 & -2.35 & 40.4 & 1000 & 4243 \\
			\hline
			\hline
			MLP-pattern &49 & 0.07 & 3.77 & 0.001 & 45 \% & 0.039 & -3.12 & 50.1 & 1000 & 1452 \\
			\hline
			MLP-vanilla &134 & 0.20 & 2.60 & 0.002 & 252 \% & 0.125 & -2.45 & 41.6 & 1000 & 3520 \\
			\hline
			MLP-candle-rep & 49 & 0.07 & 3.77 & 0.001 & 45 \% & 0.039 & -3.12 & 50.1 & 1000 & 1452 \\
			\hline
			MLP-windowed &\textbf{177} & \textbf{0.26} & 2.22 & \textbf{0.003} & \textbf{445} \% & \textbf{0.178} & \textbf{-2.19} & \textbf{38.5} & 1000 & \textbf{5457} \\
			\hline
			\hline
			CNN1D &132 & 0.19 & 2.61 & 0.002 & 244 \% & 0.123 & -2.46 & 41.7 & 1000 & 3444 \\
			\hline
			CNN2D &144 & 0.21 & 2.59 & 0.002 & 287 \% & 0.135 & -2.43 & 41.5 & 1000 & 3876 \\
			\hline
			\hline
			GRU & 166 & 0.25 & \textbf{2.43} & 0.002 & 388 \% & 0.161 & -2.31 & 40.2 & 1000 & 4888 \\	
			\hline
		\end{tabular}
	\end{adjustbox} 
\end{table*}

\begin{table*}[htb]
	\centering
	\caption{Performance of different models on AAPL.}
	\label{tbl:AAPL}
	\begin{adjustbox}{width=\textwidth}
		\begin{tabular}{@{} |l*{10}{|l}| @{}}
			\hline
			Agent & \rotatebox[origin=c]{90}{Arithmetic Return} & \rotatebox[origin=c]{90}{Average Daily Return} & \rotatebox[origin=c]{90}{Daily Return Variance} & \rotatebox[origin=c]{90}{Time Weighted Return} & \rotatebox[origin=c]{90}{Total Return}  & \rotatebox[origin=c]{90}{Sharpe Ratio} & \rotatebox[origin=c]{90}{Value At Risk} & \rotatebox[origin=c]{90}{Volatility} & \rotatebox[origin=c]{90}{Initial Investment} & \rotatebox[origin=c]{90}{Final Portfolio Value}\\
			\hline
			\hline
			Rule-Based & 74 & 0.11 & \textbf{2.11} & 0.001 & 86 \% & 0.072 & \textbf{-2.28} & \textbf{37.5} & 1000 & 1869 \\
			\hline
			B\&H & 123 & 0.18 & 4.71 & 0.002 & 191 \% & 0.085 & -3.39 & 56.0 & 1000 & 2919 \\
			\hline
			RL      & 113 & 0.17 & 4.49 & 0.001 & 162 \% & 0.079 & -3.32 & 54.7 & 1000 & 2626 \\
			\hline
			\hline
			DQN-pattern & 123 & 0.18 & 4.71 & 0.002 & 194 \% & 0.086 & -3.38 & 56.0 & 1000 & 2946 \\
			\hline
			DQN-vanilla & 165 & 0.24 & 3.06 & 0.002 & 372 \% & 0.142 & -2.63 & 45.1 & 1000 & 4722 \\
			\hline
			DQN-candle-rep & 123 & 0.18 & 4.71 & 0.002 & 192 \% & 0.085 & -3.39 & 56.0 & 1000 & 2923 \\
			\hline
			DQN-windowed & 161 & 0.24 & 3.79 & 0.002 & 341 \% & 0.124 & -2.96 & 50.2 & 1000 & 4410 \\
			\hline
			\hline
			MLP-pattern &118 & 0.17 & 4.70 & 0.002 & 179 \% & 0.082 & -3.39 & 55.9 & 1000 & 2795 \\
			\hline
			MLP-vanilla &164 & 0.24 & 3.15 & 0.002 & 365 \% & 0.139 & -2.68 & 45.8 & 1000 & 4657 \\
			\hline
			MLP-candle-rep & 123 & 0.18 & 4.71 & 0.002 & 192 \% & 0.085 & -3.39 & 56.0 & 1000 & 2923 \\
			\hline
			MLP-windowed &\textbf{184} & \textbf{0.27} & 3.52 & \textbf{0.003} & 462 \% & 0.148 & -2.81 & 48.4 & 1000 & 5623 \\
			\hline
			\hline
			CNN1D &166 & 0.25 & 3.06 & 0.002 & 376 \% & 0.143 & -2.63 & 45.1 & 1000 & 4760 \\
			\hline
			CNN2D &184 & \textbf{0.27} & 3.43 & \textbf{0.003} & \textbf{462} \% & \textbf{0.150} & -2.77 & 47.7 & 1000 & \textbf{5623} \\
			\hline
			\hline
			GRU & 174 & 0.26 & 3.27 & 0.002 & 412 \% & 0.145 & -2.71 & 46.6 & 1000 & 5129 \\	
			\hline
		\end{tabular}
	\end{adjustbox} 
\end{table*}

\begin{table*}[htb]
	\centering
	\caption{Performance of different models on KSS.}
	\label{tbl:KSS}
	\begin{adjustbox}{width=\textwidth}
		\begin{tabular}{@{} |l*{10}{|l}| @{}}
			\hline
			Agent & \rotatebox[origin=c]{90}{Arithmetic Return} & \rotatebox[origin=c]{90}{Average Daily Return} & \rotatebox[origin=c]{90}{Daily Return Variance} & \rotatebox[origin=c]{90}{Time Weighted Return} & \rotatebox[origin=c]{90}{Total Return}  & \rotatebox[origin=c]{90}{Sharpe Ratio} & \rotatebox[origin=c]{90}{Value At Risk} & \rotatebox[origin=c]{90}{Volatility} & \rotatebox[origin=c]{90}{Initial Investment} & \rotatebox[origin=c]{90}{Final Portfolio Value}\\
			\hline
			\hline
			Rule-Based & 74 & 0.11 & \textbf{2.11} & 0.001 & 86 \% & 0.072 & \textbf{-2.28} & \textbf{37} & 1000 & 1869 \\
			\hline
			B\&H & 123 & 0.18 & 4.71 & 0.002 & 191 \% & 0.085 & -3.39 & 56 & 1000 & 2919 \\
			\hline
			RL      & -15 & -0.02 & 4.71 & -0.001 & -33 \% & -0.017 & -3.59 & 56 & 1000 & 664 \\
			\hline
			\hline
			DQN-pattern & 31 & 0.04 & 12.74 & -0.000 & -9 \% & 0.013 & -5.83 & 92 & 1000 & 900 \\
			\hline
			DQN-vanilla & 272 & 0.40 & 9.24 & \textbf{0.004} & 1023 \% & 0.134 & -4.59 & 78 & 1000 & 11236 \\
			\hline
			DQN-candle-rep & -53 & -0.08 & 4.08 & -0.001 & -49 \% & -0.040 & -3.40 & 52 & 1000 & 505 \\
			\hline
			DQN-windowed & 179 & 0.27 & 10.12 & 0.002 & 332 \% & 0.085 & -4.97 & 82 & 1000 & 4328 \\
			\hline
			\hline
			MLP-pattern &106 & 0.16 & 6.74 & 0.001 & 133 \% & 0.062 & -4.11 & 67 & 1000 & 2331 \\
			\hline
			MLP-vanilla &\textbf{286} & \textbf{0.43} & 9.13 & \textbf{0.004} & \textbf{1204} \% & \textbf{0.143} & -4.54 & 78 & 1000 & \textbf{13048} \\
			\hline
			MLP-candle-rep &-42 & -0.06 & 15.63 & -0.001 & -61 \% & -0.016 & -6.57 & 102 & 1000 & 389 \\
			\hline
			MLP-windowed &170 & 0.25 & 9.55 & 0.002 & 303 \% & 0.083 & -4.83 & 79 & 1000 & 4032 \\
			\hline
			\hline
			CNN1D &278 & 0.41 & 9.35 & \textbf{0.004} & 1093 \% & 0.137 & -4.61 & 78 & 1000 & 11932 \\
			\hline
			CNN2D &173 & 0.260 & 9.537 & 0.002 & 313.508 \% & 0.084 & -4.82 & 79 & 1000 & 4135 \\
			\hline
			\hline
			GRU &175 & 0.264 & 9.746 & 0.002 & 320.412 \% & 0.085 & -4.87 & 80 & 1000 & 4204 \\	
			\hline
		\end{tabular}
	\end{adjustbox} 
\end{table*}

\subsubsection{Input types}
\label{sec:input-types}
\begin{figure*}
	\centering
	\begin{subfigure}{\textwidth}
		\includegraphics[width=\linewidth, height=0.20\textheight]{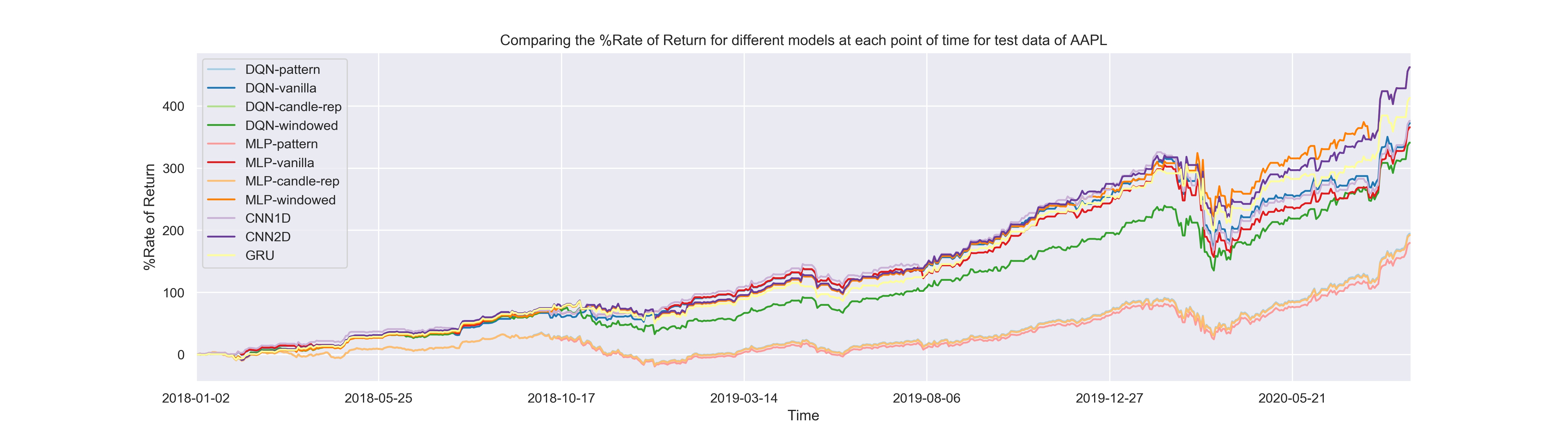}
		\caption{Performance of different models on AAPL.}
	\end{subfigure}
	\\
	\begin{subfigure}{\textwidth}
		\includegraphics[width=\linewidth, height=0.20\textheight]{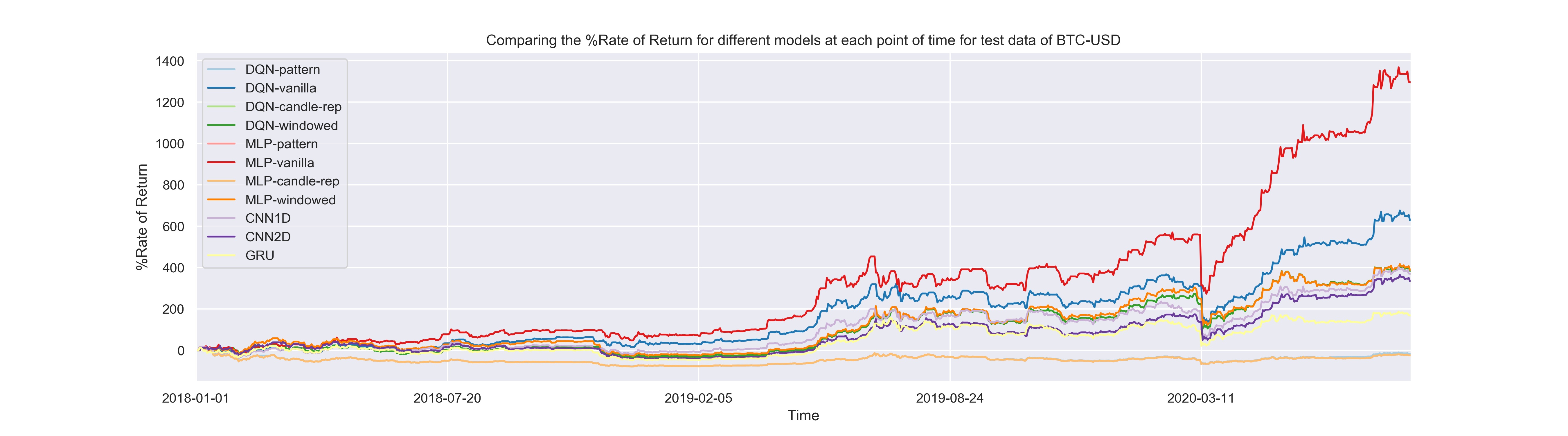}
		\caption{Performance of different models on BTC/USD pair.}
	\end{subfigure}
	\\
	\begin{subfigure}{\textwidth}
		\includegraphics[width=\linewidth, height=0.20\textheight]{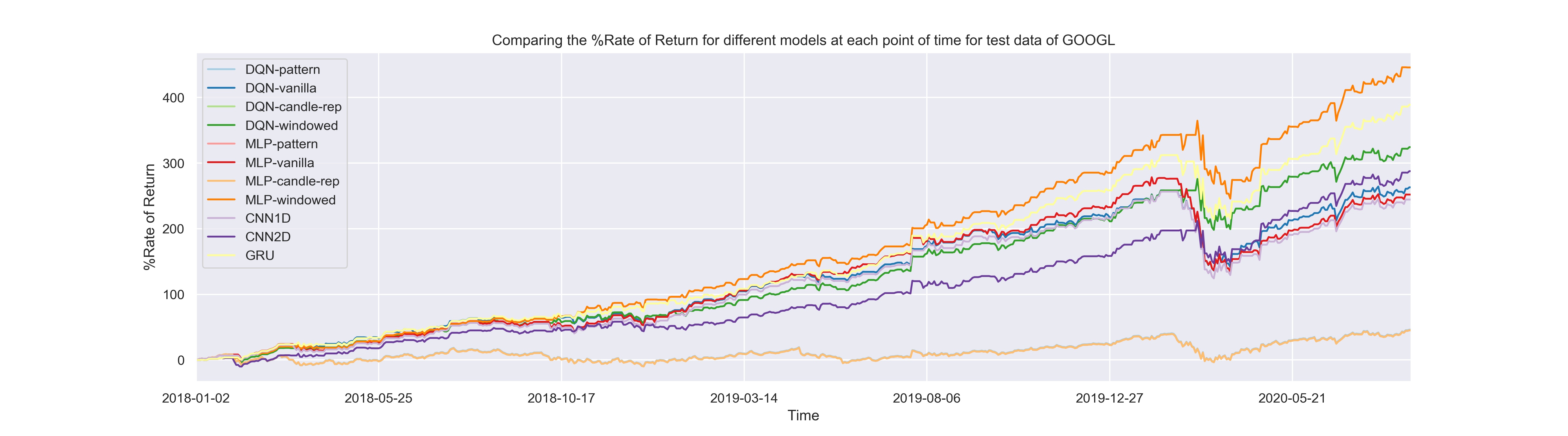}
		\caption{Performance of different models on GOOGL.}
	\end{subfigure}
	\\
	\begin{subfigure}{\textwidth}
		\includegraphics[width=\linewidth, height=0.20\textheight]{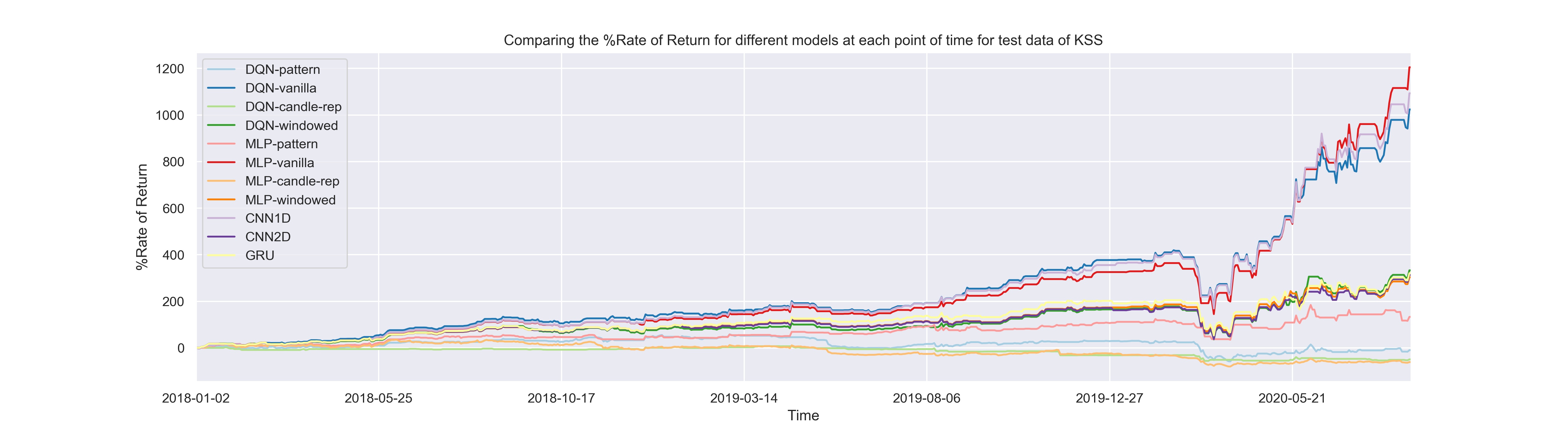}
		\caption{Performance of different models on KSS.}
	\end{subfigure}
	
	\caption{Profit curves of different models with different input types on each asset.}
	\label{fig:IPC}
\end{figure*}

Different input types provide various representations from the same information. In this experiment, we investigated the effect of using different input representations on the performance of the proposed models. The following four types of input representations are used and figure \ref{fig:IPC} illustrates the results of each model with different input types on each dataset.

\begin{enumerate}[i)]
	\item Pattern: A binary vector representing the appearance of each introduced popular candlestick patterns.
	\item Vanilla: The raw OHLC prices.
	\item Candle-rep: The representation of a candlestick's parts including a quadruple <$u$, $l$, $b$, $\nu$> denoting the percentages of the upper shadow $u$, the lower shadow $l$, the body $b$, and the direction of the candlestick $\nu$ (bullish or bearish).
	\item Windowed: The time-series of raw OHLC prices including the only last 3-time steps. 
\end{enumerate}

Since, in the popular candlestick patterns the longest sequential pattern consists of three time-steps on average (like Morning Star pattern), the inputs of the models are provided only from the last 3 time-steps. In the case that the input is provided from the last 3 time-steps, three feature extraction models are used: 1) A CNN model with 1-dimensional kernels along side the time axis (CNN1D), 2) a CNN model with 2-dimensional kernels computing an average over the spatial and temporal features (CNN2D), and 3) a GRU model to extract temporal relationships into a single feature vector. 

The performance of the time-series models on the assets in which the testing data follows the training data trends (GOOGL and AAPL) is obviously better than the other models. While, in the other datasets, the raw OHLC representation outperforms the other representations.

We expected that the performance of the models that use temporal relationships of the input time-series be better than the performance of the models using just the last time-step inputs; however, the experimental results obviously indicate that the models working only on the input of the last time-step perform better than the others. 

We believe that the short-time price fluctuations makes the decision making for the models difficult. In order to improve the performance of the models the length of the input time-series should be large enough to illustrate a trend in the price history of the asset. Furthermore, typically the frequency of appearance of the 3-step candlestick patterns in the input time-series is notably lower than the other patterns and the models cannot see enough data to learn appropriate rules for such cases. In summary, this observation illustrates that working on learning single asset trading signals should be limited to either only the last time-step data or a large enough time-series indicating at least a price trend in the input.

\subsubsection{Learned trading rules}

\begin{figure*}
	\centering
	\begin{subfigure}{\textwidth}
		\includegraphics[width=\linewidth, height=0.20\textheight]{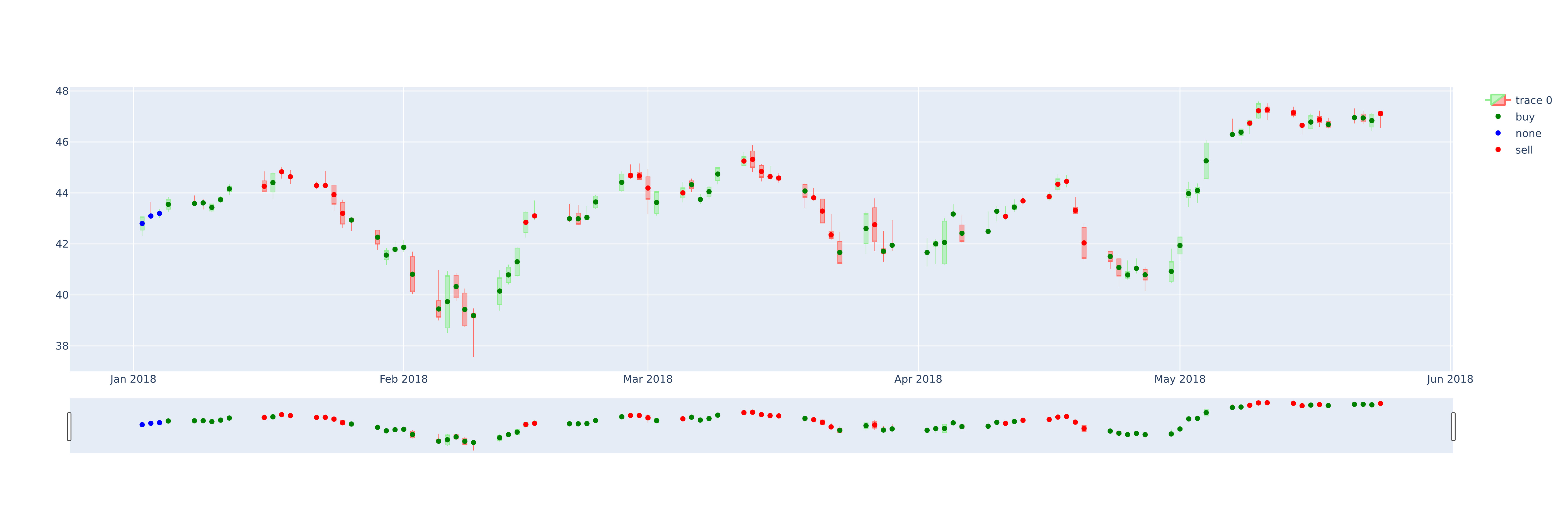}
		\caption{AAPL}
	\end{subfigure}
	\\
	\begin{subfigure}{\textwidth}
		\includegraphics[width=\linewidth, height=0.20\textheight]{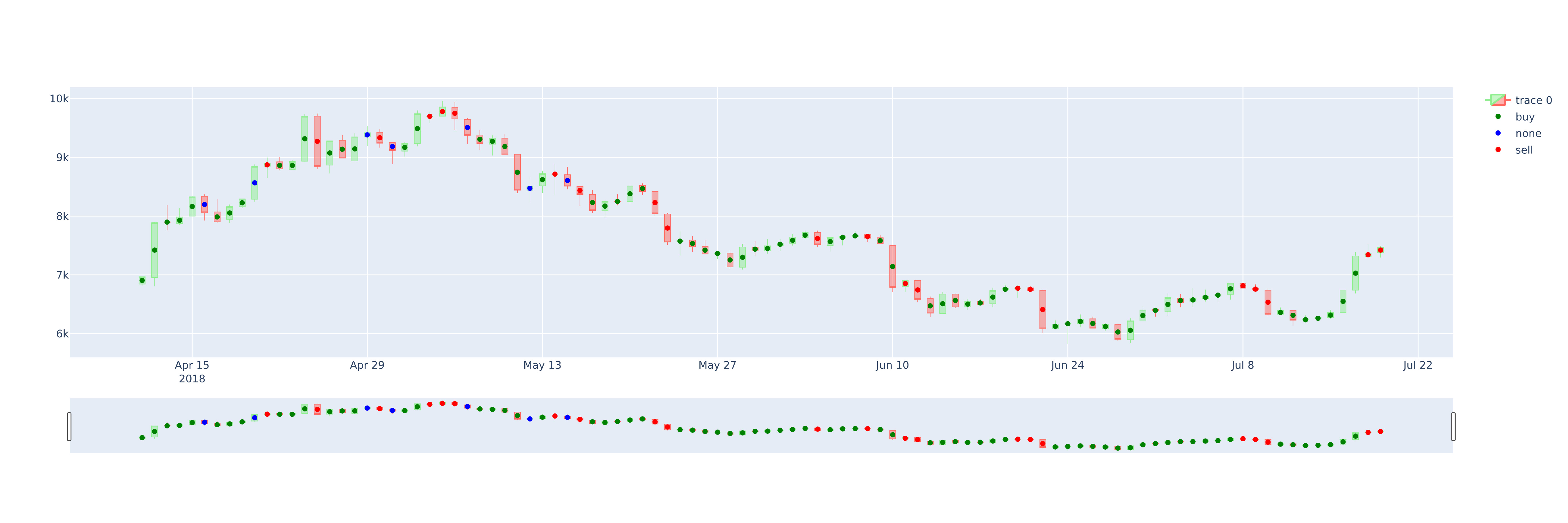}
		\caption{BTC/USD}
	\end{subfigure}
	\\
	\begin{subfigure}{\textwidth}
		\includegraphics[width=\linewidth, height=0.20\textheight]{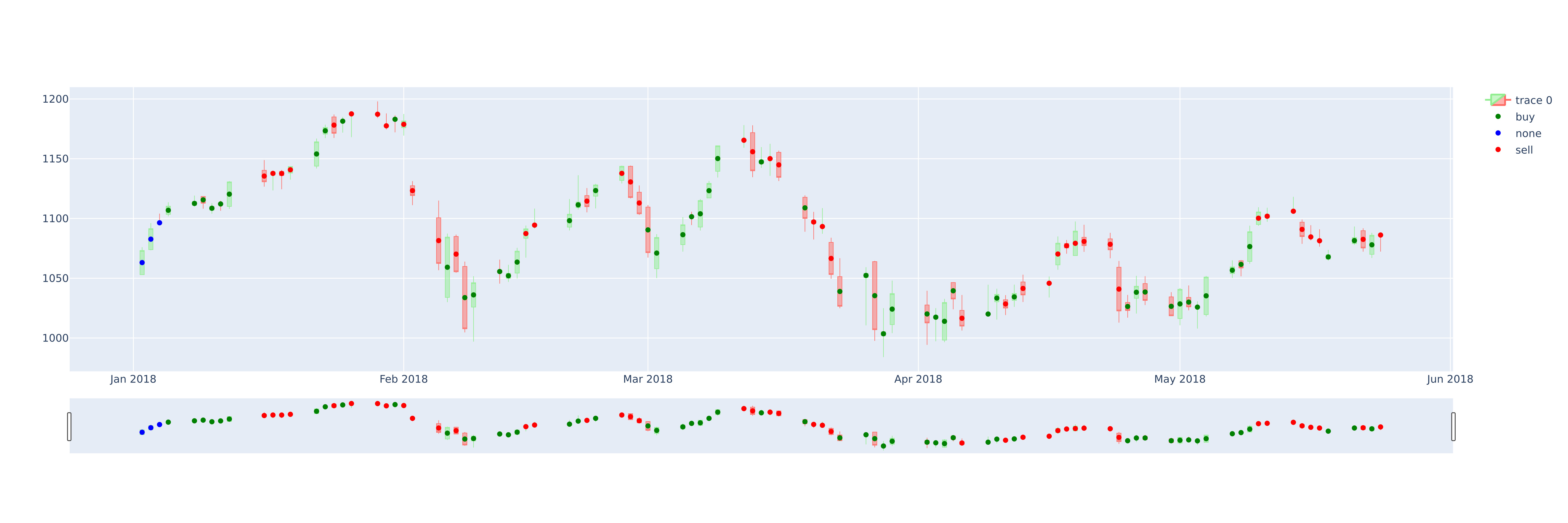}
		\caption{GOOGL}
	\end{subfigure}
	\\
	\begin{subfigure}{\textwidth}
		\includegraphics[width=\linewidth, height=0.20\textheight]{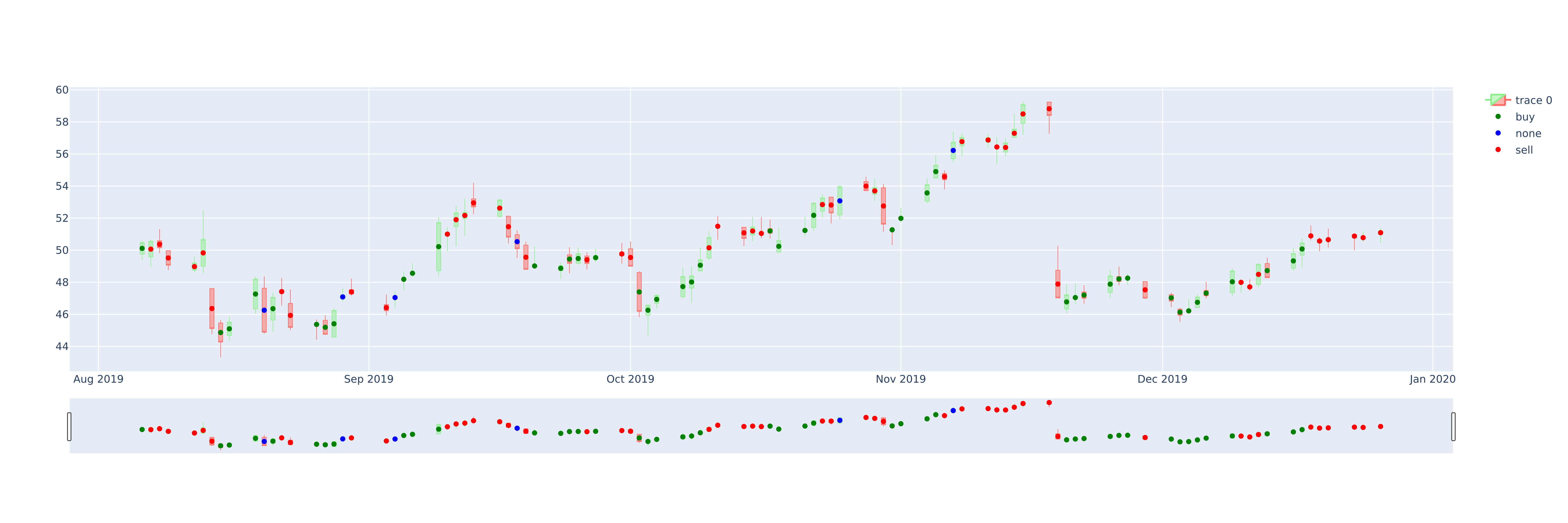}
		\caption{KSS}
	\end{subfigure}
	
	\caption{Trading decisions made by each model on each asset.}
	\label{fig:DM}
\end{figure*}

Figure \ref{fig:DM} illustrates the decisions made by the best model proposed in this paper (MLP) at each time-step for different assets. In order to get a better sense about the trading behavior of this model, only 100 decisions of this model are displayed. The green points demonstrate the 'Buy' action, the red points demonstrate the 'Sell' actions and the blue points demonstrate the 'None' action. Although the actions are displayed right at the generation point in figure \ref{fig:DM}, signals generated by the model are assumed to be accomplished on the next trading day. The model is assumed to invest all of its available money at the first time that the 'Buy' action is raised. From this point while not 'Sell' action is generated, the bought asset will not be sold. Also, when the first 'Sell' action is observed, the whole of the bought asset will be sold. So, we just evaluate the first actions and do not care about the repetitive actions. 

It seems that the learned trading rules, detect the bullish and bearish trends well and generate the 'Buy' signals almost in the first half of bullish trends, and also generate the 'Sell' signals in the second half of the bullish or bearish trends. 

\subsubsection{Comparing with similar works}
Table \ref{tbl:compare} compares the performance of our best model with the state-of-the-art models using the profit metrics. According to the results reported in table \ref{tbl:wang2017deep-compare}, the performance of the MLP model proposed in this work is significantly better than DQT and RRL on stocks HSI and S\&P500 proposed by Wang et al. \cite{wang2017deep}.

Table \ref{tbl:wu2020adaptive-compare} represents the Rate of Return (\%) of MLP model and models proposed by Wu et al. \cite{wu2020adaptive}. Wu et al.'s best performance is on AAPL stock with Rate of Return equals 77.7, but the MLP model gains the Rate of Return 262.3, which is greatly better. Moreover, wherever the models proposed by Wu et al. got a negative return, our model returns a highly positive return, e.g., on stock GE.

Table \ref{tbl:xiong2018practical-compare} shows the performance of DDPG, the model presented by Xiong et.al. \cite{xiong2018practical}. The final portfolio value of MLP is better than DDPG, starting with an initial portfolio value of 10000.

As the results indicate, the MLP model performs significantly better than similar models in profitability. As we said before, these papers' codes were not available, and we had to compare the performance according to common metrics.      

\begin{table}[htb]
	\centering
	\caption{Compare performance with similar works according to the profit}
	\begin{adjustbox}{width=\textwidth}
	\begin{subtable}[t]{0.38\textwidth}
		\begin{tabular}{|c|c|c|}
			\hline
			Agents &‌HSI & S\&P500\\
			\hline
			MLP & 13231.2 & 5032.3 \\
			\hline
			\hline
			DQT & 350 & 214\\
			\hline
			RRL & 174 & 141\\
			\hline
		\end{tabular}
		\caption{Compare profitability performance with Wang et. al. \cite{wang2017deep} based on Accumulated Return(\%)}
		\label{tbl:wang2017deep-compare}	
	\end{subtable}
\hspace{1cm}
	\begin{subtable}[t]{0.7\textwidth}
		\begin{tabular}{|c|c|c|c|c|c|}
			\hline
			Agents & AAPL & GE &‌AXP &‌CSCO & IBM\\
			\hline
			MLP & 262.3 & 130.4 & 260.2 & 259.9 & 149.2\\
			\hline
			\hline
			GDQN & 77.7 & -10.8 & 20.0 & 20.6 & 4.63\\
			\hline
			GDPG & 82.0 & -6.39 & 24.3 & 13.6 & 2.55\\
			\hline
			Turtle & 69.5 & -17.0 & 25.6 & -1.41 & -11.7\\
			\hline
		\end{tabular}
		\caption{Compare profitability performance with Wu et. al. \cite{wu2020adaptive} based on Rate of Return(\%)}
		\label{tbl:wu2020adaptive-compare}
	\end{subtable}
	\hspace{0.5cm}
	\begin{subtable}[t]{0.3\textwidth}
		\begin{tabular}{|c|c|}
			\hline
			Agents & DJI\\
			\hline
			MLP & 21580\\
			\hline
			\hline
			DDPG & 19791\\
			\hline
			Min-Variance & 14369\\
			\hline
			DJIA & 15428\\
			\hline
		\end{tabular}
		\caption{Compare profitability performance with Xiong et. al. \cite{xiong2018practical} based on Final Portfolio Value(Initial Portfolio Value is 10000)}
		\label{tbl:xiong2018practical-compare}
	\end{subtable}
	\end{adjustbox}
	\label{tbl:compare}
\end{table}

\section{Conclusion}
In this paper, we investigated the performance of deep reinforcement learning models in learning financial asset-specific trading rules and strategies. We proposed a novel deep reinforcement learning model based on the Q-learning technique to generate trading signals given different types of input representations. In addition, four feature extraction models were proposed to improve the DRL module performance on decision making: 1) An MLP network consisting of two fully-connected layers, 2) A convolutional model with one-dimensional kernels applying to OHLC prices, 3) A convolutional model with two-dimensional kernels applying to the last 3 time steps price histories, and 3) A GRU model applying to the last 3 price data. Furthermore, the performance of different models in learning trading rules for different financial assets in various situations were assessed. 

Experiments carried out on the performance of the DRL-based models demonstrate the following results:

\begin{enumerate}
	\item Learning asset-specific trading rules profits more than general experimental rules on different assets. The deep reinforcement models outperform the traditional RL methods, benchmark strategies, and general experimental rules in this task.
	\item The MLP feature extractor performs more profitable than the other feature extraction methods in cases that the price behavior of the training data is highly different than the testing data.
	\item The Convolutional and GRU feature extractors are potentially able to learn more profitable trading rules when the price behavior of training and testing sets are similar.
	\item In different input representations the short-length price time-series are not good input representations for the DRL models. The raw OHLC prices are more suitable to extract asset-specific trading rules for the deep models than the candlestick charts, or popular experimental candlestick patterns.
\end{enumerate}

This study, can be further improved in the future research in different ways. First, the performance of the time-series based models and feature extractors should be investigated. Second, since the learned trading rules of the models should be changed in testing time when a clear price behavior change is detected, the behavior change detection and rule changing abilities of such models should be investigated. Third, based on the experimental results, it is clear to the authors that proposing more complex feature extraction models, should inevitably improve the models performance.

\section*{References}

\bibliography{references}

\appendix
\section{Rules by the traditional rule-based agent}
\label{sec:app1}
In this section, we briefly discuss the details of the rule-based pattern extraction for candlestick charts, and also trading signals for each rule. These patterns along with their rules of extraction and their shapes are represented in \eqref{tbl:rules}. For each rule, we need to use some predefined hyper-parameters to specify them accurately. The value of these hyper-parameters are chosen according to the $\% Rate\;of\;Return$ of the rule-based method. 

\begin{itemize}
	\item GSL(Gap Significance Level):
	This is a number between 0 and 1, showing the significance level of the gap (difference between the opening and closing price of 2 consecutive candles). It is multiplied by the length of the candle with largest body of the two candles whose gap length needs to be evaluated.
	
	\item CSL(Candle Significance Level):
	This is a number between 0 and 1, showing if a candle length is significant enough in order to be part of a pattern. This number is multiplied by the maximum-body-length candle in a dataset. Then, the length of the candles in a specific pattern is compared with the max-body-length candle to see if its length is significant enough(not too small).
	
	\item PSH(Percentage of Shadow Hammer):
	This parameter shows that the upper shadow of the hammer should be at most what percentage of the total length.
	
	\item UBHL(Upper Bound Hammer Length) \& LBHL(Lower Bound Hammer Length):
	These two parameters show the boundaries that the body of the hammer should be inside them in order to be considered a hammer.	
\end{itemize}

We need to declare some functions for simple calculations on the candle sticks. These functions are defined in \eqref{tbl:functions}. 
For each pattern, to produce correct signals, the trend of the stock before that pattern occurs should be calculated. In \eqref{eq:signals} $\psi_t(P_t)$
shows the trading signal at time step $t$. This function is used in $Signaling Function$ of \eqref{tbl:rules}. In \eqref{eq:signals}, MT is defined in \eqref{eq:trend}.
\begin{equation}
\psi_t(P_t) = \begin{cases}
\text{'Buy'} &\quad\text{if MT is downtrend}\\
\text{'None',} &\quad\text{if MT is side}\\
\text{'Sell'} &\quad\text{otherwise}\\
\end{cases}
\label{eq:signals}
\end{equation}

\begin{table*}
	\centering
	\caption{A list of defined function used in the process of pattern extraction}
	\begin{adjustbox}{width=\textwidth}
	\begin{tabular}{|c|c|c|c|Sc|}
		\hline
		Number& Function Name  & Definition & Return Value & Candle Shape
		\\
		\hline
		\hline
		1& $IsBull(P)$ & Is candle P bullish or not‌& $
		\begin{cases}
		\text{'True'} &\quad\text{if } P_{close} > P_{open} \\
		\text{'False'} &\quad\text{otherwise}\\
		\end{cases}
		$ & \includegraphics[height=1.5cm]{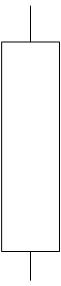}
		\\
		\hline
		2& $IsBear(P)$ & Is candle P bearish or not & $
		\begin{cases}
		\text{'True'} &\quad\text{if } P_{open} > P_{close} \\
		\text{'False'} &\quad\text{otherwise}\\
		\end{cases}
		$ & \includegraphics[height=1.5cm]{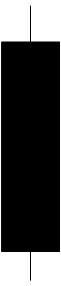}
		\\
		\hline
		3& $TL(P)$ & Total length of candle P & $P_{high} - P_{low}$ & \includegraphics[height=1.5cm]{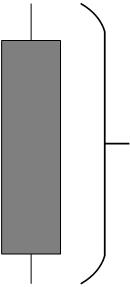}
		\\
		\hline
		4& $BL(P)$ & Body length of candle P & $|P_{close} - P_{open}|$ & \includegraphics[height=1.5cm]{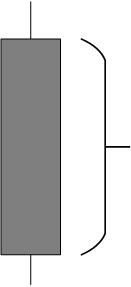}
		\\
		\hline
		5& $USL(P)$ & Upper shadow length of candle P & $
		\begin{cases}
		P_{high} - P_{close} &\quad\text{if } IsBull(P) \\
		P_{high} - P_{open} &\quad\text{if } IsBear(P)\\
		\end{cases}
		$ & \includegraphics[height=1.5cm]{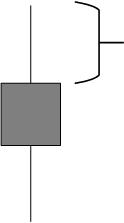}
		\\
		\hline
		6& $LSL(P)$ & Lower shadow length of candle P & $
		\begin{cases}
		P_{open} - P_{low} &\quad\text{if } IsBull(P) \\
		P_{close} - P_{low} &\quad\text{if } IsBear(P)\\
		\end{cases}
		$ & \includegraphics[height=1.5cm]{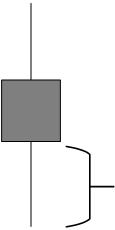}
		\\
		\hline
		7& $IsLS(P)$ & Is length of P significant & $BL(P) \geq CSL \times \max_{i \in [0 \dots n]}P_i$ & \includegraphics[height=1.5cm]{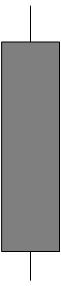}
		\\
		\hline
		8& $MidPoint(P)$ & Middle point of P & $\frac{P_{close} + P_{open}}{2}$ & \includegraphics[height=1.5cm]{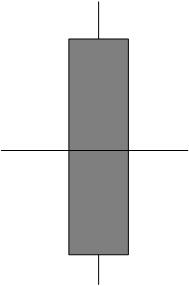}
		\\
		\hline
		9& $IsDoji(P)$ & Is P doji & $
		\begin{cases}
		\text{'True'} &\quad\text{if } P_{close} \approx P_{open} \\
		\text{'False'} &\quad\text{otherwise}\\
		\end{cases}
		$ & \includegraphics[height=1cm]{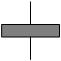}
		\\
		\hline
		10& $GS(P_1, P_2)$ & Gap significance & $GSL \times max(BL(P_1), BL(P_2)))$ & \includegraphics[height=1.5cm]{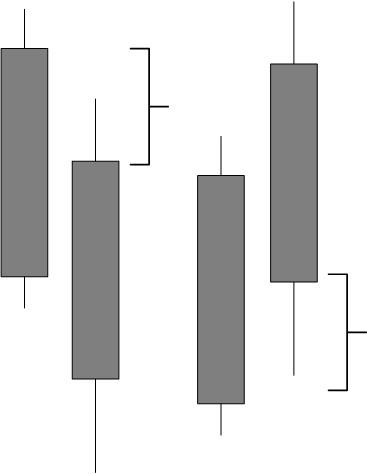}
		\\
		\hline
	\end{tabular}
	\end{adjustbox}
	\label{tbl:functions}
\end{table*}

\begin{table*}
	\centering
	\caption{A list of used trading rules based on popular candlestick patterns \cite{keller2006candlestick}}
	\begin{adjustbox}{width=\textwidth}
	\begin{tabular}{|c|c|c|Sc|c|}
		\hline
		Number& Name  & Definition & Chart & Signaling Function
		\\
		\hline
		\hline
		1& Hammer / Inverse Hammer & 
		\parbox{3cm}{\begin{align*}
			&IsBull(P);\\
			&\begin{cases}
			(P_{high} - P_{close}) \leq PSH \times TL(P); &\quad\text{If P is Hammer}\\
			(P_{open} - P_{low}) \leq PSH \times TL(P); &\quad\text{If P is Inverse Hammer}
			\end{cases}\\
			&LBHL \times TL(P) \leq BL(P) \leq UBHL \times TL(P);  
			\end{align*}} 
		& \includegraphics[height=1.5cm]{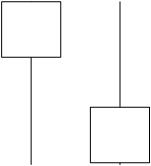} & buy if downtrend 
		\\
		\hline
		2& Hanging Man / Shooting Star & 
		\parbox{3cm}{\begin{align*}
			&IsBear(P);\\
			&\begin{cases}
			(P_{high} - P_{open}) \leq PSH \times TL(P); &\quad\text{If P is Hanging man}\\
			(P_{close} - P_{low}) \leq PSH \times TL(P); &\quad\text{If P is Shooting star}
			\end{cases}\\
			&LBHL \times TL(P) \leq BL(P) \leq UBHL \times TL(P);  
			\end{align*}} 
		& \includegraphics[height=1.5cm]{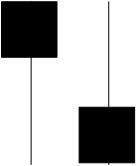} & sell in uptrend 
		\\
		\hline
		3& Bullish Engulfing &  
		\parbox{3cm}{\begin{align*}
			&IsLS(P_2) \\
			&P_{2, open} \leq P_{1, close} \leq P_{2, close}\\
			&P_{2, open} \leq P_{1, open} \leq P_{2, close} 
			\end{align*}} & \includegraphics[height=1.5cm]{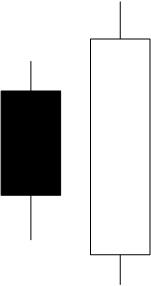} & buy if downtrend
		\\
		\hline
		4& Bearish Engulfing &  
		\parbox{3cm}{\begin{align*}
			&IsLS(P_2) \\
			&P_{2, close} \leq P_{1, close} \leq P_{2, open}\\
			&P_{2, close} \leq P_{1, open} \leq P_{2, open} 
			\end{align*}} & \includegraphics[height=1.5cm]{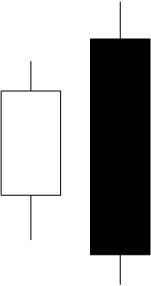} & sell if uptrend
		\\
		\hline
		5& Bullish Harami &  
		\parbox{3cm}{\begin{align*}
			&IsLS(P_1);\;IsBear(P_1);\;IsBull(P_2) \\
			&P_{2,close} \leq P_{1,open}\\
			&P_{2,open} - P_{1, close} \geq GSL \times BL(P_1)
			\end{align*}}  & \includegraphics[height=1.5cm]{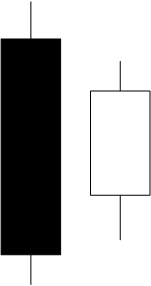} & buy if downtrend
		\\
		\hline
		6& Bearish Harami &  
		\parbox{3cm}{\begin{align*}
			&IsLS(P_1);\;IsBull(P_1);\;IsBear(P_2) \\
			&P_{2,close} \geq P_{1,open}\\
			&P_{1, close} - P_{2,open}\geq GSL \times BL(P_1)
			\end{align*}}  & \includegraphics[height=1.5cm]{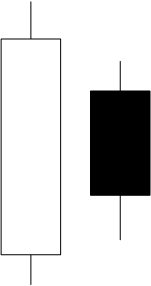} & sell if uptrend
		\\
		\hline
		7& Piercing Line &   
		\parbox{3cm}{\begin{align*}
			&IsLS(P_1);\;IsLS(P_2);\;IsBear(P_1);\;IsBull(P_2) \\
			&GS(P_1, P_2) \leq (P_{1, close} - P_{2, open}) \\
			&(P_{2, close} \geq MidPoint(P_1))
			\end{align*}} & \includegraphics[height=1.5cm]{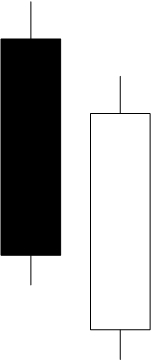} & buy if downtrend
		\\
		\hline
		8& Dark Cloud Cover &    
		\parbox{3cm}{\begin{align*}
			&IsLS(P_1);\;IsLS(P_2);\;IsBull(P_1);\;IsBear(P_2) \\
			&GS(P_1, P_2) \leq (P_{2, open} - P_{1, close}) \\
			&(P_{2, close} \leq MidPoint(P_1))
			\end{align*}} & \includegraphics[height=1.5cm]{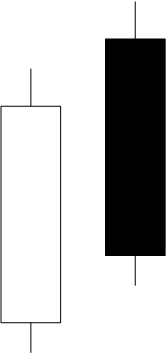} & sell if uptrend
		\\
		\hline
		9& Morning Star &    
		\parbox{3cm}{\begin{align*}
			&IsLS(P_1);\;IsLS(P_3);\;IsBear(P_1);\;IsDoji(P_2);\;IsBull(P_3) \\
			&(P_{2, close} \leq P_{3, open}) \\
			&(P_{2, close} \leq P_{1, close}) 
			\end{align*}} & \includegraphics[height=1.5cm]{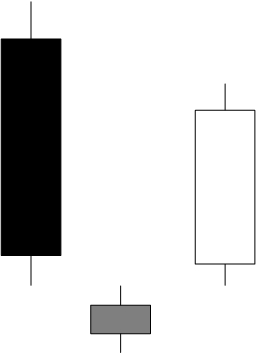} & buy if downtrend
		\\
		\hline
		10& Evening Star &    
		\parbox{3cm}{\begin{align*}
			&IsLS(P_1);\;IsLS(P_3);\;IsBull(P_1);\;IsDoji(P_2);\;IsBear(P_3) \\
			&(P_{2, close} \geq P_{3, open}) \\
			&(P_{2, close} \geq P_{1, close}) 
			\end{align*}} & \includegraphics[height=1.5cm]{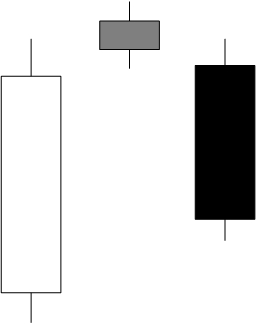} & sell if uptrend
		\\
		\hline	
	\end{tabular}
	\end{adjustbox}
	\label{tbl:rules}
\end{table*}

\begin{table*}
	\centering
	\begin{adjustbox}{width=\textwidth}
	\begin{tabular}{|c|c|c|Sc|c|}
		\hline
		Number& Name  & Definition & Chart & Signaling Function
		\\
		\hline
		\hline
		11& Three White Soldiers &     
		\parbox{3cm}{\begin{align*}
			&IsLS(P_1);\;IsLS(P_2);\;IsLS(P_3)\\
			&IsBull(P_1);\;IsBull(P_2);\;IsBull(P_3) 
			\end{align*}} & \includegraphics[height=1.5cm]{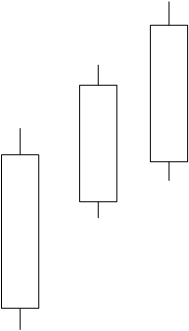} & buy if downtrend
		\\
		\hline
		12& Three Black Crows &     
		\parbox{3cm}{\begin{align*}
			&IsLS(P_1);\;IsLS(P_2);\;IsLS(P_3)\\
			&IsBear(P_1);\;IsBear(P_2);\;IsBear(P_3) 
			\end{align*}} & \includegraphics[height=1.5cm]{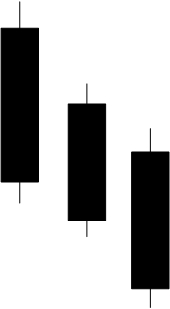} & sell if uptrend
		\\
		\hline
		13& Rising Three Methods &    
		\parbox{3cm}{\begin{align*}
			&IsBull(P_1);\;IsBull(P_5);\;IsBear(P_2);\;IsBear(P_3);\;IsBear(P_4)\\
			&IsLS(P_1);\;IsLS(P_2);\;IsLS(P_3);\;IsLS(P_4);\;IsLS(P_5)\\
			&max(P_{2,open},P_{3,open},P_{4,open}) \leq P_{5,high}\\
			&min(P_{2,close},P_{3,close},P_{4,close}) \geq P_{1,low}\\ 
			\end{align*}} & \includegraphics[height=1.5cm]{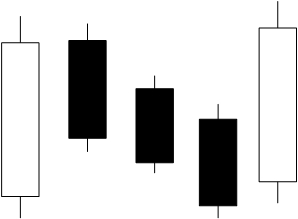} & none (indecision)
		\\
		\hline
		14& Falling Three Methods &    
		\parbox{3cm}{\begin{align*}
			&IsBear(P_1);\;IsBear(P_5);\;IsBull(P_2);\;IsBull(P_3);\;IsBull(P_4)\\
			&IsLS(P_1);\;IsLS(P_2);\;IsLS(P_3);\;IsLS(P_4);\;IsLS(P_5)\\
			&max(P_{2,close},P_{3,close},P_{4,close}) \leq P_{5,high}\\
			&min(P_{2,open},P_{3,open},P_{4,open}) \geq P_{1,low}\\ 
			\end{align*}}  & \includegraphics[height=1.5cm]{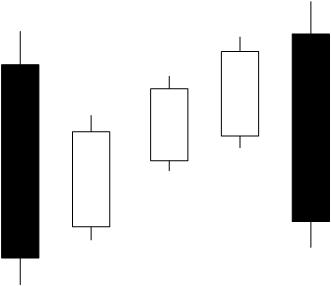} & none (indecision)
		\\
		\hline	
	\end{tabular}
	\end{adjustbox}
\label{tbl:rules2}
\end{table*}

\end{document}